%% file: main.tex
\documentclass[dvipsnames]{article} %
\usepackage{colm2024_conference}

\usepackage{booktabs}
\usepackage{graphicx}
\usepackage{enumitem}
\usepackage{wrapfig}
\usepackage{algorithm}
\usepackage{algpseudocode}

\usepackage{afterpage}
\usepackage{microtype}
\usepackage{amsmath}
\usepackage{colortbl}
\usepackage[utf8]{inputenc}
\definecolor{lightgray}{rgb}{0.9,0.9,0.9}
\usepackage{caption}
\usepackage{subcaption}
\usepackage{setspace}
\usepackage{url}
\usepackage{multirow}
\usepackage{colortbl}
\usepackage{tabularx}
\usepackage{blindtext}
\usepackage{pgfplots}
\pgfplotsset{compat=1.18} 
\usepackage{tikz}
\usetikzlibrary{er,positioning,bayesnet}
\usepackage{makecell}
\usepackage{tipa}
\usepackage{siunitx}
\usepackage{nicefrac}
\usepackage{tocloft}
\usepackage{listings}
\usepackage[raster,skins]{tcolorbox} %
\usepackage{xltabular}
\usepackage{colortbl}
\usepackage{adjustbox}
\usepackage{xurl}
\usepackage{rotating}
\usepackage[normalem]{ulem}
\useunder{\uline}{\ul}{}
\usepackage{crossreftools}

\usepackage{amsthm}
\usepackage{amssymb}
\usepackage{mathtools}
\usepackage{amsmath}
\usepackage{float}
\usepackage{graphicx} 
\usepackage[capitalize,noabbrev,nameinlink]{cleveref}
\crefname{assumption}{Assumption}{Assumptions}

\theoremstyle{plain}

\crefname{proposition}{proposition}{propositions}
\Crefname{proposition}{Proposition}{Propositions}

\theoremstyle{definition}

\theoremstyle{remark}

\makeatletter
\def\maketag@@@#1{\hbox{\m@th\normalfont\normalsize#1}}
\makeatother

\makeatletter
\pretocmd{\contentsline}
  {\begingroup\hypersetup{linkcolor=black}}{}{}
\apptocmd{\contentsline}
  {\endgroup}{}{}
\makeatother

\input{math_commands.tex}

\renewcommand{\ghlink}{https://github.com/limix-ldm/LimiX/}

\newcommand*\justify{%
  \fontdimen2\font=0.4em
  \fontdimen3\font=0.2em
  \fontdimen4\font=0.1em
  \fontdimen7\font=0.1em
  \hyphenchar\font=`\-
}

\renewcommand{\texttt}[1]{%
  \begingroup
  \ttfamily
  \begingroup\lccode`~=`/\lowercase{\endgroup\def~}{/\discretionary{}{}{}}%
  \begingroup\lccode`~=`[\lowercase{\endgroup\def~}{[\discretionary{}{}{}}%
  \begingroup\lccode`~=`.\lowercase{\endgroup\def~}{.\discretionary{}{}{}}%
  \catcode`/=\active\catcode`[=\active\catcode`.=\active
  \justify\scantokens{#1\noexpand}%
  \endgroup
}

\title{LimiX-2: A Contextual Mechanism Network Towards General Structured-Data Intelligence}

\author{
{\normalsize \bf{LimiX Team}} \\
\vspace{10pt}
 \rm Stable AI \& Tsinghua University
}

\begin{document}
\maketitle

\vspace{1em}

\begin{abstract}

We introduce LimiX-2, a new model in the LimiX family, developed through model and data scaling guided by our previously established scaling laws. LimiX-2 adopts the Contextual Mechanism Networks (CMNs) paradigm and is pretrained with Context-Conditional Masked Modeling (CCMM). CMNs shifts the organizing principle of in-context learning from target-centric prediction to mechanism-oriented joint modeling. Rather than centering the network on the $p(y \mid x, D_{\mathrm{context}})$ objective of conventional tabular PFNs, it is designed around learning $p(x, y \mid D_{\mathrm{context}})$, a context-dependent representation of the joint structure underlying data generation. Pretraining uses synthetic datasets generated by structural causal models (SCMs) spanning diverse graph structures, functional mechanisms, and observation processes. Evaluations on TabArena, TALENT, and BCCO show that LimiX-2 outperforms current dataset-specific models and tabular foundation models.
Beyond predictive performance, the CMN paradigm also promotes causal awareness in LimiX-2: its feature attention encodes direct causal relationships, enabling accurate causal skeleton recovery.

\vspace{-0pt}
\end{abstract}

\begin{figure}[h]
    \centering
    \includegraphics[width=\linewidth]{figures/evaluation/all3benchmark_1106.pdf}
    \caption{\textbf{Performance overview on evaluated benchmarks.} \ourA~achieves Elo scores of $1963$, $1510$, and $1439$ on TabArena~\citep{erickson2025tabarena}, TALENT~\citep{ye2024closerlookdeeplearning, liu2024talenttabularanalyticslearning}, and BCCO~\citep{limix2025limix}, respectively, outperforming all compared foundation models and AutoGluon 1.6~\citep{erickson2020autogluon}. For TabArena, only the best AutoGluon setting is shown. Bar ends and labels indicate Elo point estimates, while shaded regions with terminal markers show 95\% bootstrap confidence intervals.}
    \label{fig:all_leaderboard}
\end{figure}

\clearpage
\setcounter{tocdepth}{2} 
\tableofcontents
\clearpage
\input{sections/intro}
\input{sections/model_and_pretrain}

\input{sections/data_generation}

\input{sections/experiments}

\input{sections/scaling_law} 

\input{sections/conclusion}

\clearpage
\input{sections/contribution}

\clearpage

\bibliography{biblio}
\bibliographystyle{colm2024_conference}

\clearpage
\appendix

\end{document}

%% file: math_commands.tex
\usepackage{amsmath,amsfonts,bm}

\def\1{\bm{1}}

\DeclareMathAlphabet{\mathsfit}{\encodingdefault}{\sfdefault}{m}{sl}
\SetMathAlphabet{\mathsfit}{bold}{\encodingdefault}{\sfdefault}{bx}{n}

\newcommand{\ourA}{LimiX-2}

%% file: sections/intro.tex
\section{Introduction}

Progress toward general-purpose machine intelligence can be organized
around three complementary frontiers: language, the physical world,
and structured data~\citep{limix2025limix}.
Large-scale next-token pretraining and post-training have enabled
large language models (LLMs) to follow instructions, use tools, and
reason over text and visual inputs~\citep{ouyang2022training,
schick2023toolformer,team2023gemini,guo2025deepseek}.
Embodied agents and world models pursue physical-world intelligence
by learning to model and interact with environments~\citep{ha2018world,
kim2024openvla}.
In contrast, general-purpose learning and reasoning over
structured data remain comparatively underdeveloped.

Structured data supports prediction and decision-making in healthcare~\citep{johnson2016mimic}, finance~\citep{gu2020empirical}, and scientific discovery~\citep{baldi2014searching}.
For tabular prediction, gradient-boosted trees~\citep{chen2016xgboost,ke2017lightgbm,dorogush2018catboost}, deep neural networks~\citep{gorishniy2021revisiting,gorishniy2024tabr}, and automated ensemble pipelines~\citep{erickson2020autogluon} have achieved strong task-specific performance.
However, these methods typically require separate training and model selection for each dataset, with limited reuse of knowledge across tasks.
This motivates foundation models that learn from diverse datasets through pretraining and transfer to new prediction tasks.

Motivated by this gap, we advocate the development of \emph{large structured-data models} (LDMs)\footnote{Throughout this report, LDMs refers to large structured-data models.}: large-scale pretrained models for inference over structured data. We characterize LDMs by three properties: (i) pretraining on large-scale data that covers a wide distribution of tasks; (ii) a unified modeling paradigm over structured variables without task-specific design; and (iii) the ability to perform new tasks at inference time without any model updates. Classification and regression are the canonical instances of tabular data inference, yet they do not exhaust the choice of such tasks.

One line of work pretrains tabular predictors on synthetic tasks using Prior-Data Fitted Networks (PFNs)~\citep{muller2021transformers, hollmann2022tabpfn,hollmann2025tabpfn}.
Subsequent models, including TabICL~\citep{qu2025tabicl,jingangtabiclv2}, TabFM~\citep{kong2026tabfm}, and Mitra~\citep{zhang2025mitra,tao2026mitrav2}, follow this approach of pretraining on synthetic tasks for supervised in-context prediction. In their standard supervised formulation, PFNs approximate $p(y \mid x, D_{\mathrm{context}})$, the posterior predictive distribution of a designated target $y$ given query features $x$ and a labeled context set~\citep{hollmann2022tabpfn}.
This enables prediction on new datasets through in-context learning without parameter updates.
However, within each task, direct supervision is confined to predicting a single target column rather than explicitly modeling the joint distribution over all variables, which may significantly limit the ability on variant data reasoning tasks.

We introduce \emph{Contextual Mechanism Networks} (CMNs), a new paradigm for structured-data intelligence that shifts the modeling focus from a designated target to the system of predictive dependencies among variables.
Unlike PFNs, CMNs learn from multiple conditional prediction tasks over the same dataset, using context to capture the dependencies shared across them and target the joint dependency structure of
$p(\mathbf{x}, y \mid D_{\mathrm{context}})$.
Supervised prediction is thus a special case of a broader framework for inferring unobserved quantities from available evidence.

This paradigm builds on the context-conditional modeling principles introduced in LimiX~\citep{limix2025limix}.
We instantiate CMNs in LimiX-2 and advance this design through model and data scaling. Pretraining uses \emph{Context-Conditional Masked Modeling} (CCMM)~\citep{limix2025limix}, which integrates target prediction and feature reconstruction under varied observation patterns.
By organizing supervision across variables, CCMM makes inter-variable inference an explicit pretraining objective rather than an auxiliary capability.
This provides a unified basis for supervised prediction, missing-value imputation, and broader conditional reasoning within a single pretrained model.

LimiX-2 refines the Transformer-based architecture of LimiX~\citep{vaswani2017attention,limix2025limix} while retaining cell-level representations.
It is pretrained exclusively on synthetic datasets produced by an expanded generation engine based on structural causal models (SCMs)~\citep{pearl2003causality}. Compared with its predecessor, the engine spans a broader range of graph structures, functional mechanisms, and observation processes.
Our evaluations demonstrate the effectiveness of the CMN paradigm: a single pretrained LimiX-2 model supports classification, regression, and missing-value imputation, as well as causal discovery, without task-specific parameter updates.

We empirically evaluate the prediction performance of LimiX-2 model on three  benchmarks widely adopted by the community of tabular machine learning: TabArena, TALENT and BCCO. These benchmarks span broad regimes of sample size, feature dimensionality, class number, categorical–to-numerical feature ratio, missingness and sample-to-feature ratios. The results demonstrate that our LimiX-2 model outperforms current models, including tabular foundation models and traditional models trained specifically on each dataset. Notably, LimiX-2 surpasses TabFM despite being 4 times smaller in model parameter size. 
Furthermore, we also conduct causal skeleton recovery evaluations on several typical causal discovery datasets. The results demonstrate that LimiX-2 outperforms other tabular foundation models, tree-based feature importance methods and dedicated causal discovery methods, indicating that its feature attention encodes causal structural information.

%% file: sections/model_and_pretrain.tex
\section{Architecture} \label{sect:architecture}

\ourA~continues the cell-level design of the previous generation~\citep{limix2025limix}: it does not compress feature information at the row level, instead encodes each cell into separate representation\footnote{For brevity, the terms representation and embedding are used interchangeably throughout this report and refer to the same concept.}, supporting conditional reasoning across variables. Local relations among features within a row can therefore be modeled directly, while dataset-level statistics can be derived from the corresponding cells in context. Contrastive to representations that collapse a whole row into a single vector, cell-level representations better preserve the fine-grained structure tabular data.

On this basis, \ourA~instantiates cell-level modeling at a larger scale. Every cell must keep its own representation, and masked prediction further requires these vectors to yield consistent conditionals under different visibility patterns.
The parameter budget is increased so that each cell has a richer representation space and fine-grained relations across columns and samples can be captured more stably. The extra capacity is not applied as a uniform width multiplier. It is allocated mainly to the subsequent task pathway, which organizes the evidence needed for prediction. 

\subsection{Embedding of Tabular Data}
\label{sec:embedding}
Suppose a table has $N$ rows and $F$ columns, we denote raw cell in the $i$-th row and $j$-th column as $x_{i,j}^{R}$ and raw targets of the $i$-th row as $y_{i}^{R}$. We firstly map each raw cell $x_{i,j}^{R}$ into feature representation space $\mathbf{x}_{i,j} \in \mathbb{R}^{d}$. 
In the previous version LimiX, we set $d=192$ for LimiX-16M~\citep{limix2025limix} and $d=96$ for LimiX-2M~\citep{wang2026limix}. \ourA~ extends the embedding dimension to $d=256$.  
Missing cells share a single learnable embedding, while column identity is provided separately by discriminative feature encoding (DFE).

$$
\mathbf{x}_{i,j}=
\begin{cases}
E_{\mathrm{miss}}, & \text{if } x_{i,j}^{R} \text{ is missing}, \\
E_{\mathrm{num}}(x_{i,j}^{R}), & \text{otherwise}.
\end{cases}
$$

$E_{\mathrm{num}}$ is a two-layer MLP with hidden dimensions of 128 and 256, respectively. Each linear layer is followed by RMSNorm\citep{zhang2019rmsnorm} and ReLU activation.  Therefore, the entire feature representation tensor can be denoted as $\mathbf{x}\in\mathbb{R}^{N\times F\times d}$.

Targets are encoded as $\mathbf{y}_i \in \mathbb{R}^{Kd}$ (we set $K=4$ in LimiX-2) according to task type: numerical regression targets are mapped through an encoder $E^{Y}_{\mathrm{num}}$, while categorical classification targets are mapped through an orthogonally initialized embedding table $E^{Y}_{\mathrm{cat}}$. Context rows retain their observed labels, whereas the target position of each query row is filled with a learnable \texttt{MASK} embedding. While LimiX uses the task embedding as a whole per sample, \ourA~splits the embedding into $K$ task-embedding slots, each of dimension $d$. Formally,
\[
\mathbf{y}_{i}
=
\bigl(\mathbf{y}_{i,1}, \ldots, \mathbf{y}_{i,K}\bigr),
\qquad
\mathbf{y}_{i,k} \in \mathbb{R}^{d}.
\]

Furthermore, a task-type embedding $E_{\mathrm{type}}(\tau)$, where $\tau \in \{\mathrm{cls}, \mathrm{reg}\}$, is added to each of the four task embeddings.

\subsection{Discriminative Feature Encoding}

In \ourA, each feature shares the same numerical MLP, so the distinct columns of similar marginal distribution become  indistinguishable from their value representations alone. Therefore, an explicit column identity is therefore required.

\ourA~adopts low-rank DFE to produce column identity representation. The $j$-th column is associated with an $s$-dimensional code $u_{j} \in \mathbb{R}^{s}$, where $s=d/4$ by default. A transformation matrix $E\in\mathbb{R}^{s\times d}$ then maps the codes into the embedding space, and the mapped codes (i.e. column identity embedding) are added with feature representation. 

The column identity embedding $e_{j}$ distinguish the columns without encoding sequential proximity: when columns are permuted together with their codes, attention should not depend on an accidental order. The low rank confines column identity to a compact subspace so that statistical strength can be shared across columns. Since the representation dimension is set as a larger value $d=256$ than previous generation of LimiX, it becomes easier to memorize column-index shortcuts. Hence, compressing column identity into an $s$-dimensional code constrains the model recognizing columns rather than positions.

\subsection{Model Backbone Architecture}

\begin{figure}[t]
    \centering
    \includegraphics[width=\linewidth]{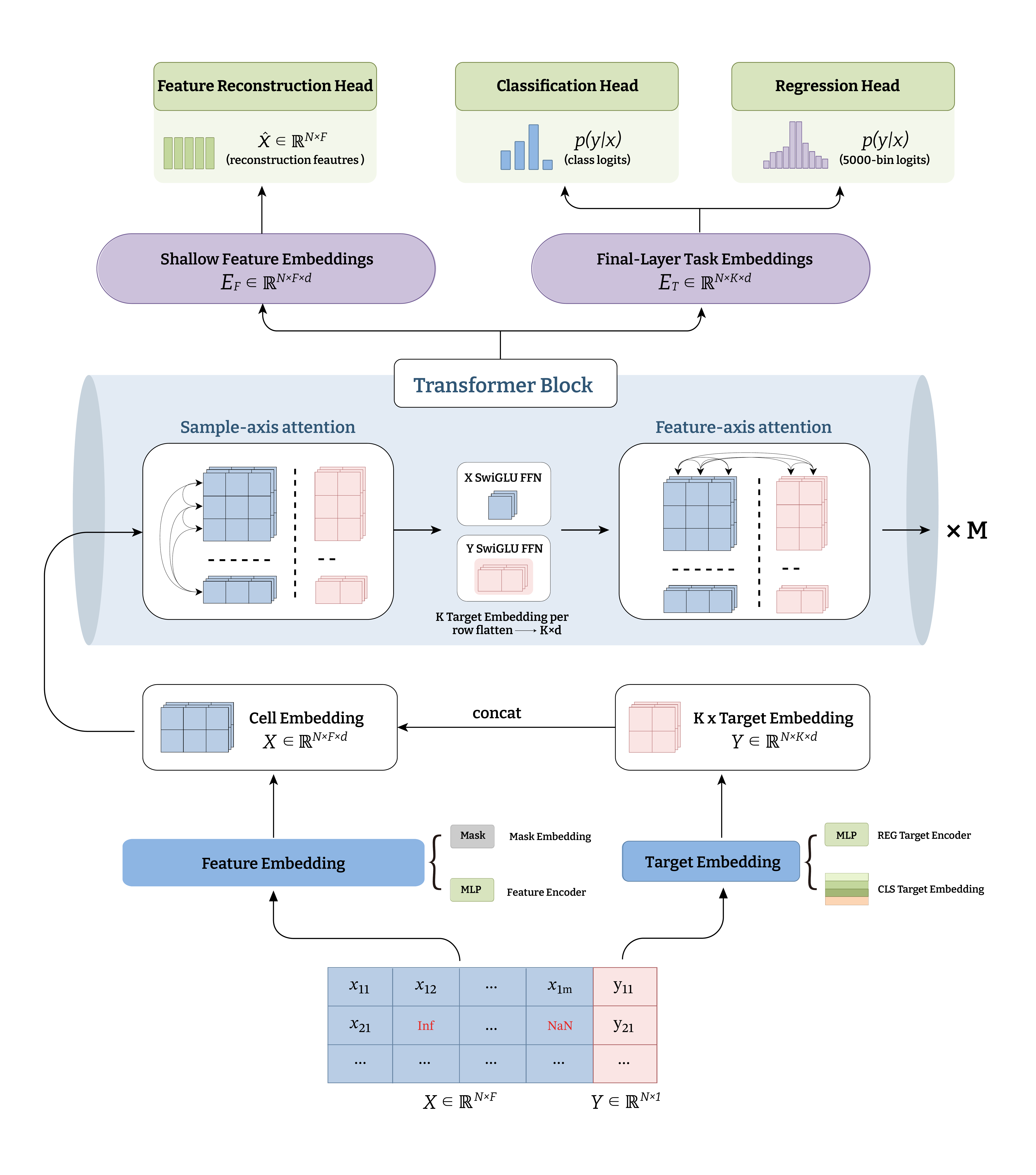}
    \caption{\textbf{Overall structure of \ourA.} Features are encoded by a MLP-based encoder, and missing cell is imputed by a shared learnable vector. Targets are encoded into $K=4$ embedding slots. Each block successively applies sample-axis attention and SwiGLU on the feature and target representations, followed by asymmetric feature-axis attention. Feature embeddings produced at the shallow depth are used for masked-feature reconstruction. Final-layer task embeddings are used for both classification and regression tasks.}
    \label{fig:limix_v2_dataflow}
\end{figure}

The backbone remains a stack of dual-axis transformer blocks: the feature-axis blocks models variable relations within rows, and the sample-axis blocks use context samples to form a prior for the current table. In LimiX, feature-axis and sample-axis attention and the Feed-Forward Network (FFN) were shared within a block~\citep{limix2025limix}. \ourA~keeps this two-dimensional factorization, but no longer routes feature and task representations through the same computation path. The model architecture is shown in \Cref{fig:limix_v2_dataflow}. We denote $\mathbf{x}_{i,j}^{(l)}$ and $\mathbf{y}^{(l)}$ as the output representation of the $l$-th dual-axis transformer block. Specifically, $\mathbf{x}_{i,j}^{(0)}$ and $\mathbf{y}^{(0)}_{i}$ are the original representation of the initial embedding components. Formally, we have $\mathbf{x}_{i,j}^{(0)}=\mathbf{x}_{i,j}$ and $\mathbf{y}^{(0)}_{i}=\mathbf{y}_{i}$.

\paragraph{Sample-axis attention.}  For each dual-axis transformer block, the representations produced by the previous block are firstly fed into sample-axis attention components. The sample-axis attention components propagate information across samples on each feature position as well as target position. For the target position, LimiX-2 firstly concatenates the task embeddings into a unified one before feeding into attention process,

$$
\mathbf{y}_{i}^{(l-1)}=(\mathbf{y}_{i,1}^{(l-1)},\ldots,\mathbf{y}_{i,K}^{(l-1)})\in\mathbb{R}^{Kd}.
$$

In the attention component, context rows are visible to each other, while query rows can only attend to context. The query/key/value mapping functions are shared among features, but not between features and target. A query prediction therefore depends only on the sample’s own features and the context, not on which other test samples share the batch. After the calculation, the target representaions are then split back into $K$ embeddings. The immediate representation of features and target produced by the $l^{th}$ sample-axis attention are denoted as $\widetilde{x}_{i,j}^{(l)}$ and $\widetilde{y}_{i}^{(l)}$ respectively.

\paragraph{Asymmetric feature-axis attention.} Feature representations may attend to the target and other feature  representations, while target representations can only attend to features representations. Formally we have,

$$
\mathbf{x}^{(l)}=\mathrm{Attn}_{X}^{\mathrm{feat}}\big(Q_{X}(\widetilde{\mathbf{x}}^{(l)}),\,K_{X}([\widetilde{\mathbf{x}}^{(l)},\widetilde{\mathbf{y}}^{(l)}]),\,V_{X}([\widetilde{\mathbf{x}}^{(l)},\widetilde{\mathbf{y}}^{(l)}])\big),
$$

$$
\mathbf{y}^{(l)}=\mathrm{Attn}_{Y}^{\mathrm{feat}}\big(Q_{Y}(\widetilde{\mathbf{y}}^{(l)}),\,K_{Y}(\widetilde{\mathbf{x}}^{(l)}),\,V_{Y}(\widetilde{\mathbf{x}}^{(l)})\big),
$$
where $\mathrm{Attn}_{X/Y}^{\mathrm{feat}}(\cdot,\cdot,\cdot)$ is multi-head attention components for feature/target representations and $Q_{X/Y}(\cdot)$, $K_{X/Y}(\cdot)$ and $V_{X/Y}(\cdot)$ are query mapping function, key mapping function and value mapping function of feature and target representation attention respectively.
The Q/K/V mapping functions for feature and target representation attention are not shared, so the model implicitly distinguish their roles inside a mixed stream.

\paragraph{Independent SwiGLU.} The shared MLP is replaced by a gated FFN~\citep{shazeer2020glu}, instantiated separately for feature and target representations. Formally,

$$
\mathrm{SwiGLU}(z)=W_{o}\big[\mathrm{SiLU}(W_{g}z+b_{g})\odot(W_{v}z+b_{v})\big]+b_{o},
$$
where $W_{g/v}$ and $b_{g/v/o}$ are learnable projection matrices and bias parameters.

The FFNs of feature representations operates in the space of $\mathbb{R}^{p}$, and the FFNs of target representations operates on the concatenated slot space of $\mathbb{R}^{Kd}$. 

\paragraph{Multi-head Attention and length stability of multi-head attention.} In LimiX, cross-attention used only the one key/value head. In contrast, \ourA~applies all the K/V heads. Before attention scores are computed, $Q$ and $K$ are normalized to control the magnitude of the attention logits in deep stacks. Queries are then rescaled per head by a length-dependent factor

$$
s_{h}=(1+w_{h}\log n)\,\beta_{h},
$$

where $n$ is the current sequence length and $w_{h}$ and $\beta_{h}$ are learnable and softly truncated by the tanh function. This is conceptually related to length-aware softmax scaling for variable context lengths~\citep{chiang2022overcoming,nakanishi2025scalable}. All sublayers use pre-normalized RMSNorm~\citep{zhang2019rmsnorm}.

The computation order inside a block is as follows: independent $X$/$Y$ sample-axis attention, independent SwiGLU, asymmetric feature-axis attention, and residual connections. The blocks are stacked $M=24$ layers deep.

\subsection{Prediction Heads}
\label{sec:pred_heads}
In LimiX-2, the prediction heads of different tasks (i.e. classification, regression and masked-feature reconstruction) are attached to the output representation of different depths. Masked-feature reconstruction needs local details of data and the corresponding prediction head is attached to the shallow depth representations $\mathbf{x}^{(l_{\mathrm{mask}})}$ ($l_{\mathrm{mask}}<M$). In contrast, classification and regression tasks are decoded from the representations of the last layer $\mathbf{y}^{(M)}$. Each head is preceded by an independent bottleneck post-adapter (Post Adapter): $A_{\mathrm{mask}}$, $A_{\mathrm{cls}}$, or $A_{\mathrm{reg}}$. For classification and regression, the post-adapter is applied to each of the $K$ target  embeddings slots, which are then concatenated into the space of $\mathbb{R}^{Kd}$.

For $C$-way classification, the head emits logits in $\mathbb{R}^{C}$ and is trained with cross-entropy. Learning objective of regression does not adopt mean square error (MSE) loss as in LimiX. Instead, \ourA~partitions the target range into $B=5000$ ordered bins, predicts probability of  each bin $p\in\Delta^{B-1}$, and derive the regression value as following

$$
\hat{y}=\sum_{i=1}^{B}p_{i}\,c_{i},
$$

where $c_{i}$ is the  center value of the $i^{th}$ bin.

\section{Pretraining} \label{sect:pretraining}

\subsection{Context-Conditional Masked Modeling for Joint Distribution Learning}

Pretraining aims to capture the joint dependency structure of table
variables through conditional prediction under varied observation
patterns.
Following LimiX~\citep{limix2025limix}, \ourA~adopts
\emph{Context-Conditional Masked Modeling} (CCMM), which combines
target prediction with masked-feature reconstruction.
Whereas standard supervised objectives in PFNs including TabPFN and TabICL
concentrate on $p(y \mid \mathbf{x}, D_{\mathrm{ct}})$ within each
task~\citep{hollmann2025tabpfn,qu2025tabicl}, CCMM extends direct
supervision across variables and conditioning sets.

Each pretraining episode partitions a table into disjoint context
and query row sets, $\mathcal{I}_{\mathrm{ct}}$ and
$\mathcal{I}_{\mathrm{te}}$.
The context $D_{\mathrm{ct}} =
(\mathbf{X}_{\mathrm{ct}}, \mathbf{y}_{\mathrm{ct}})$ retains
available observations, providing evidence about the table's
marginal distributions and inter-variable dependencies.
For each query row $i \in \mathcal{I}_{\mathrm{te}}$, let $\pi_i$
index its masked feature columns.
The model estimates

$$
q_{\theta}\ \big(x_{i,j}\mid \mathbf{x}_{i,-\pi_i},\,\mathbf{X}_{\mathrm{ct}},\,\mathbf{y}_{\mathrm{ct}}\big),\qquad i\in\mathcal{I}_{\mathrm{te}},\ j\in\pi_i,
$$

where $j$ ranges over the masked columns of row $i$, and $\mathbf{x}_{i,-\pi_i}$ denotes the observed query features.
The query target $y_i$ is predicted from the same conditioning
information through the task heads.

Along the sample axis, query rows attend only to context rows,
and context representations are computed without access to queries.
This prevents both direct and context-mediated information exchange
between query rows.
For fixed input representations and context, predictions are
therefore invariant to query-batch composition.
The same conditional interface supports classification, regression,
and masked-feature reconstruction without task-specific parameter updates.
Conditional likelihoods can additionally be used to score
potentially anomalous entries.

\ourA~retains CCMM while scaling an architecture with separate
feature and task pathways.
The feature pathway models inter-variable dependencies through
cell-level representations, while the task pathway uses $K=4$
embeddings to aggregate prediction-relevant information.
Asymmetric attention directs information from feature representations
to the task readout.
We scale the backbone and widen the task FFN to provide additional
capacity for conditional prediction over wider tables, longer
contexts, and more diverse observation patterns.

\subsection{Mask Pattern Design}

A fixed masking pattern restricts the range of conditional prediction tasks encountered during pretraining. \ourA~therefore combines three masking schemes to vary the granularity of prediction targets and the available conditioning information.

Masks are applied to individual entries, selected columns across query rows, or blocks of entries, exposing the model to prediction tasks at different granularities.
The resulting tasks range from recovering isolated values to predicting target columns and reconstructing groups of missing entries, all conditioned on the remaining observations and the context set.
Interleaving these schemes across episodes broadens the coverage of observation patterns and discourages specialization to a single reconstruction setting.

\subsection{Mask Embedding}

For each cell masked during pretraining, the value embedding is replaced by the shared missing-value embedding $E_{\mathrm{miss}}$ defined in~\Cref{sec:embedding} and added to the DFE column code $e_j$.
Masked and naturally missing cells thus share a missingness encoding while retaining column identity. The resulting representations pass through the same dual-axis attention layers as observed-cell embeddings, allowing the model to integrate evidence across columns and context samples and produce distributional predictions through the corresponding output heads.

%% file: sections/data_generation.tex
\section{Pretraining Data Generation}

We construct large-scale pretraining data following the SCM framework ~\citep{pearl2003causality}. By varying the components at different stages of the data-generation process, we synthesize a large number of datasets with diverse variable dependencies, feature distributions, and task properties.

\begin{figure}[t]
    \centering
    \includegraphics[width=\linewidth]{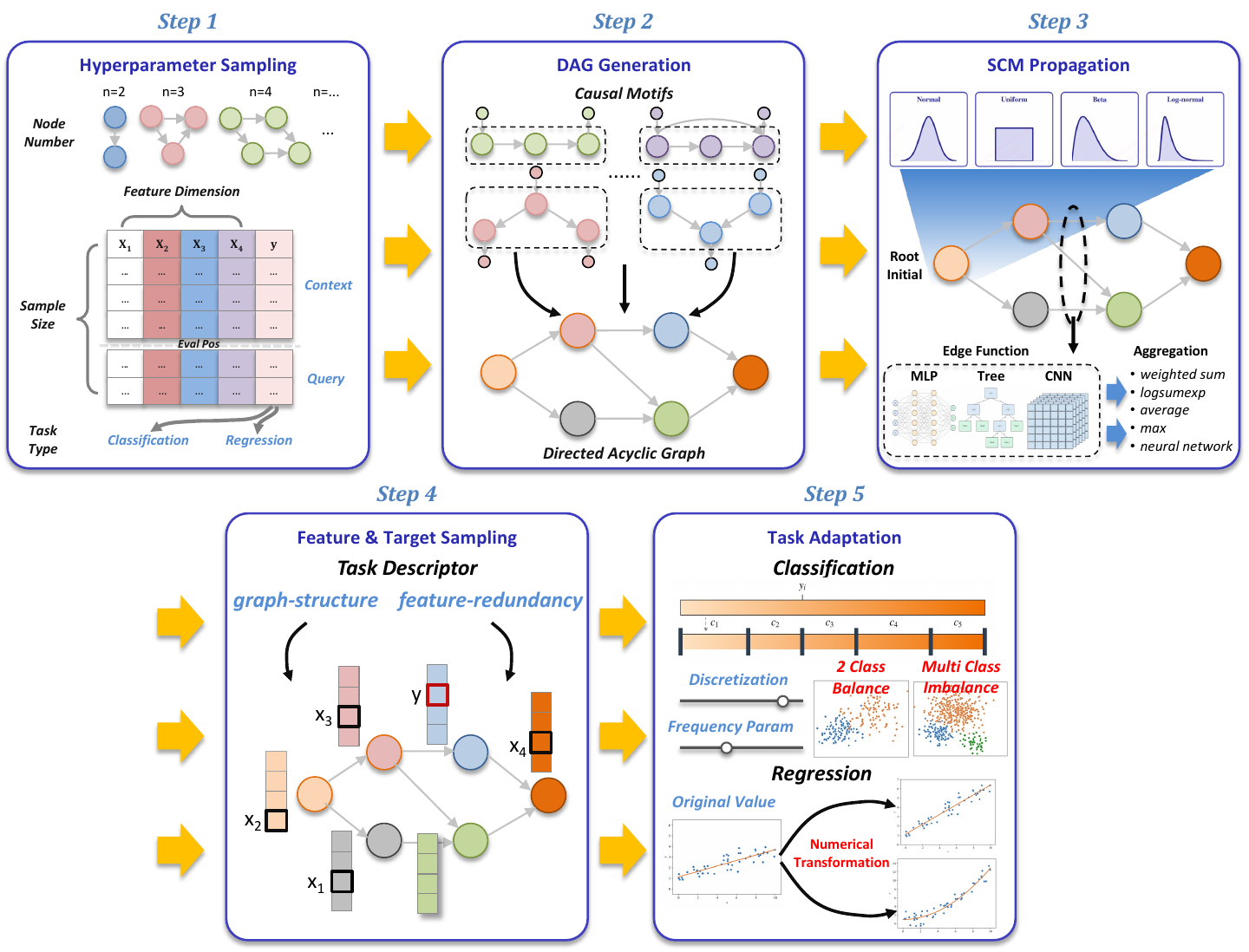}
    \caption{\textbf{ Schematic overview of the synthetic data generation process for pretraining.} The pipeline consists of five phases, that are hyperparameter sampling, directed acyclic graph (DAG) generation, SCM propagation, data sampling and task adaptation respectively.}
    \label{fig:data_generation}
\end{figure}

Inherited from the previous version of LimiX~\citep{limix2025limix}, the overall data-generation pipeline consists of five main stages-hyperparameter sampling, directed acyclic graph (DAG) generation, SCM propagation, data sampling, and task adaptation-as illustrated in~\Cref{fig:data_generation}. Building upon this pipeline, LimiX-2 further expands the space of graph structures, functional mechanisms, and variable observation processes, thereby increasing the structural and statistical diversity of the pretraining tasks.

\subsection{Hyperparameter Sampling}
For each pretraining dataset, we sample a set of hyperparameters that characterize its global properties, including the sample size, the feature dimension (specified separately as the numbers of continuous and categorical features), and the task type (i.e., classification or regression). Given the sampled sample size, we then randomly draw an evaluation position that splits the dataset into a context part and a query part. The sampling distribution of each hyperparameter is randomly chosen from a family of distributions, such as the normal, uniform, and beta distributions.

\subsection{Directed Acyclic Graph Generation}

We generate DAGs that depict the structural dependencies among variables in a hierarchical manner. The overall DAG is composed of multiple local causal structures (LCSs), which is specified as causal motifs~\citep{barjavsic2021causal} in this practice. Each causal motif may contain multiple input and output nodes and encodes directed dependencies among variables, such as chain, confounding, and collider structures~\citep{peters2017elements, pearl2016causal}.

Through recursive expansion of causal motifs at multiple granularities, the induced DAG can simultaneously capture macro- and micro-level dependencies with complex local topologies.
In addition, we allow topology-constrained graph transformation~\citep{maslov2002specificity,sanfeliu1983distance} on the DAG, where operations such as edge redirection, local path replacement, and node-level structural transformations are randomly selected and applied. These operations further enrich the local topology while preserving the acyclicity of the graph, allowing the resulting DAGs to exhibit diverse connectivity patterns and information-propagation pathways.

\subsection{Functional Mechanisms of SCM}

While the DAG of an SCM determines the dependencies among variables, the functional mechanisms determine the statistical relationships underlying these dependencies.

For each DAG, we first sample the values of root nodes from distributions with randomly chosen types and parameters. The values of the remaining nodes are then determined by propagating functional computations along the topological order of the DAG. For a node $X_i$, the value-generation process can be expressed as

$$
X_i = f_i\left(\{g_{i,j}(X_j)\}_{j\in \mathrm{PA}(X_i)}, \epsilon_i\right),
$$

where $\mathrm{PA}(X_i)$ denotes the set of parent nodes of $X_i$, $g_{i,j}$ is the edge function associated with the parent node $X_j$, $f_i$ denotes the aggregation function that combines the mapped parent values, and $\epsilon_i$ denotes stochastic noise.

LimiX-2 retains the three main types of edge functions from the previous version, including MLPs, CNNs, and decision trees~\citep{limix2025limix}. Furthermore, we incorporate additional mechanisms such as linear mappings, kernel functions, piecewise functions, periodic functions, and multiplicative interactions. Moreover, these basic functions can be composed, enabling the SCM to depict more complex variable relationships.

For nodes with multiple parents, we design several aggregation strategies, e.g., simple averaging, weighted aggregation, and neural aggregation, producing diverse forms of multivariate interactions.

\subsection{Feature and Target Sampling}

A complete SCM characterizes the joint states of all variables, whereas in practice only a subset of them is typically observable. To make the generated data resemble real-world scenarios, we retrieve only a subset of variables as sample features and prediction targets to construct each dataset. Specifically, LimiX-2 formulates variable sampling as a multi-attribute selection problem. For each dataset, coarse-grained task descriptors are constructed based on the assigned design with respect to its subgraph structure and feature redundancy. Candidate tasks are then filtered through a multi-objective selection mechanism, ensuring that the resulting tasks not only differ in graph structure but also cover prediction problems with diverse statistical properties, thereby broadening the coverage of the final pretraining tasks in both structural complexity and statistical characteristics.

\subsection{Task Adaptation}
We apply stochastic observation transformations to the features and target variables for task adaptation. Examples of such transformations include linear scaling, monotonic nonlinear transformations, logarithmic transformations, exponential transformations, and multiple operators can be randomly composed. For the target variable, the transformation is specified according to the task type of each dataset (i.e., classification or regression).
For classification tasks, since the initial target variable is generally continuous, we convert it into a categorical target via random discretization: LimiX-2 randomly partitions the value space of the target into several intervals and varies the class frequencies and discretization parameters, allowing different synthetic datasets to yield classification tasks with varying numbers of classes and degrees of class imbalance. 
For regression tasks, targets may undergo random scale transformations as well as adjustments to their skewness and tail behavior, thereby covering continuous prediction tasks with diverse functional relationships.

%% file: sections/experiments.tex
\begingroup
\newcommand{\todo}[1]{\textcolor{red}{TODO: #1}}
\newcommand{\data}[1]{\textcolor{blue}{DATA: #1}}
\newcommand{\nop}[1]{}

\definecolor{gold}{HTML}{FF0000}
\definecolor{silver}{HTML}{000000}
\definecolor{bronze}{HTML}{000000}
\definecolor{evalrow}{HTML}{EEF3F8}

\section{Evaluation}

In this section, we conduct a comprehensive predictive performance evaluation of \ourA~on several public tabular benchmarks, which encompass a diverse collection of real-world classification and regression tasks. As shown in~\Cref{tab:selected-overall,tab:talent_results,tab:bcco_results}, \ourA~achieves the highest Elo rating on all three benchmarks.
We further evaluate causal skeleton recovery with LimiX-2 on causal discovery benchmarks. The results show that the recovered skeletons based on LimiX-2 feature attention surpass those of other tabular foundation models, XGBoost-based feature importance methods, and dedicated causal discovery methods. 

\subsection{Predictive Performance Evaluation Setup}

\paragraph{Benchmarks.} 
Three widely adopted benchmarks, TabArena~\citep{erickson2025tabarena}, TALENT~\citep{ye2024closerlookdeeplearning, liu2024talenttabularanalyticslearning}, and BCCO~\citep{limix2025limix}, are used for evaluation. TabArena focuses on evaluating predictive performance under practical protocols, TALENT evaluates generalization across diverse task types, and BCCO evaluates robustness under challenging and incomplete datasets. These benchmarks provide a broad assessment of \ourA~in terms of predictive performance, generalization, scalability, and robustness.

\begin{itemize}
    \item \textbf{TabArena} is an actively maintained benchmark designed to evaluate the practical predictive performance of tabular learning methods. It comprises 51 manually curated real-world tabular datasets, covering a diverse range of classification and regression tasks. 

    \item \textbf{TALENT} is a large-scale and systematic benchmark for studying tabular learning methods across a broad spectrum of datasets. It consists of 300 datasets, including 120 binary classification datasets, 80 multiclass classification datasets, and 100 regression datasets. Excluding 12 classification datasets with more than 10 target classes, we perform evaluation on the remaining 288 datasets.

    \item \textbf{BCCO} provides a robustness-focused benchmark for tabular models, designed to evaluate model performance on challenging and imperfect data. It contains 106 classification datasets in BCCO-CLS and 50 regression datasets in BCCO-REG, focusing on scenarios involving missing and incomplete features. 
\end{itemize}

\begin{table}[t]
    \centering
    \caption{Task composition and evaluation metrics of the 3 tabular benchmarks. Twelve TALENT datasets with more than 10 target classes are excluded.}
    \label{tab:benchmark_protocol}
    \setlength{\tabcolsep}{5pt}
    \renewcommand{\arraystretch}{1.15}
    \begin{tabular}{@{}l c l c c@{}}
        \toprule
        Benchmark & \# Datasets & Task Type & \# Tasks & Main Metrics \\
        \midrule
        \multirow{3}{*}{TabArena}
            & \multirow{3}{*}{51}
            & Binary CLS     & 30  & \multirow{9}{*}{Elo, Rank, Win Rate} \\
        & & Multiclass CLS & 8   & \\
        & & REG            & 13  & \\
        \cmidrule(lr){1-4}
        \multirow{3}{*}{TALENT}
            & \multirow{3}{*}{288}
            & Binary CLS     & 120 & \\
        & & Multiclass CLS & 68  & \\
        & & REG            & 100 & \\
        \cmidrule(lr){1-4}
        \multirow{3}{*}{BCCO}
            & \multirow{3}{*}{156}
            & Binary CLS     & 71  & \\
        & & Multiclass CLS & 35  & \\
        & & REG            & 50  & \\
        \bottomrule
    \end{tabular}
\end{table}

\paragraph{Benchmark Protocol.} 
\Cref{tab:benchmark_protocol} summarizes the composition of task types in the tabular benchmarks and the main evaluation metrics reported in this technical report. The detailed evaluation protocols for TabArena, TALENT, and BCCO are described as follows.

\begin{itemize}
    \item \textbf{TabArena Protocol}: For TabArena, we fully follow the benchmark's evaluation configuration, including its dataset collection, data splits, metrics, and result aggregation procedure. We take the published leaderboard scores as the reference for existing baselines and then report the Elo rating computed by the official TabArena evaluation pipeline, ensuring direct comparability with the leaderboard results.
    
    \item \textbf{TALENT and BCCO Protocol}: For TALENT and BCCO, we follow the official evaluation pipeline provided by TALENT, conducting each experiment with 15 random seeds. For TALENT, we adopt the official fixed 64\%/16\%/20\% train/validation/test split. For BCCO, we randomly hold out 20\% of the training set for validation, leaving the official test set unchanged. 
    
    For models integrated into the TALENT library, we use their default configurations, including model-specific hyperparameter settings. Regarding tabular foundation models such as EXAONE Tabular~\citep{eo2026exaonetabular}, Xiaomi-TabLDM~\citep{wang2026xiaomitabldm}, and Mitra-v2~\citep{tao2026mitrav2}, we implement the evaluation following the preprocessing and inference procedures specified in their respective official repositories, without tuning the hyperparameters for fair comparisons.
    
    For each dataset, models are ranked by accuracy for classification tasks and by RMSE for regression tasks. Ranking is performed for each random seed, with ranks first averaged across seeds and then equally across datasets to obtain the mean rank. In addition to ranks, we aggregate model performance by converting per-dataset results into pairwise comparisons and fitting a Bradley--Terry Elo model, following TabArena, with Random Forest anchored at 1000. For reported Elo, we show 95\% confidence intervals from 2000 bootstrap rounds, using the 2.5\% and 97.5\% quantiles. 
\end{itemize}

\paragraph{Baselines.}
We compare \ourA~against a range of state-of-the-art baseline models, including tree-based models, auto-ML frameworks, neural networks, and recent tabular foundation models.

\begin{itemize}
    \item \textbf{Tree-based models.}\quad We include XGBoost~\citep{chen2016xgboost}, LightGBM~\citep{ke2017lightgbm}, CatBoost~\citep{dorogush2018catboost}, and Random Forest~\citep{breiman2001random}. 

    \item \textbf{Auto-ML frameworks.}\quad We compare \ourA~against AutoGluon~\citep{erickson2020autogluon}, an automated machine learning framework that performs model selection, hyperparameter optimization, and ensemble learning for tabular data.
    
    \item \textbf{Neural networks.}\quad We evaluate \ourA~against neural baselines, including FT-Transformer \citep{gorishniy2024tabr}, RealMLP~\citep{holzmuller2024betterrealmlp}, TabR~\citep{gorishniy2024tabr}, TabM~\citep{gorishniy2025tabm}, and ModernNCA~\citep{ye2024revisitingmodernnca}. 
 
    \item \textbf{Tabular foundation models.}\quad We compare \ourA~with recent tabular foundation models, including Mitra-v2~\citep{tao2026mitrav2}, TabFM~\citep{kong2026tabfm}, TabPFN-3~\citep{grinsztajn2026tabpfn3}, TabICLv2~\citep{jingangtabiclv2}, TabDPT~\citep{ma2026tabdpt}, EXAONE Tabular~\citep{eo2026exaonetabular}, Xiaomi-TabLDM~\citep{wang2026xiaomitabldm} and LimiX-16M~\citep{limix2025limix}. These models cover a range of recent approaches to pretraining and in-context learning for tabular data.
\end{itemize}

\subsection{Results on TabArena}

\begin{figure}[t]
    \centering
    \includegraphics[width=\linewidth]{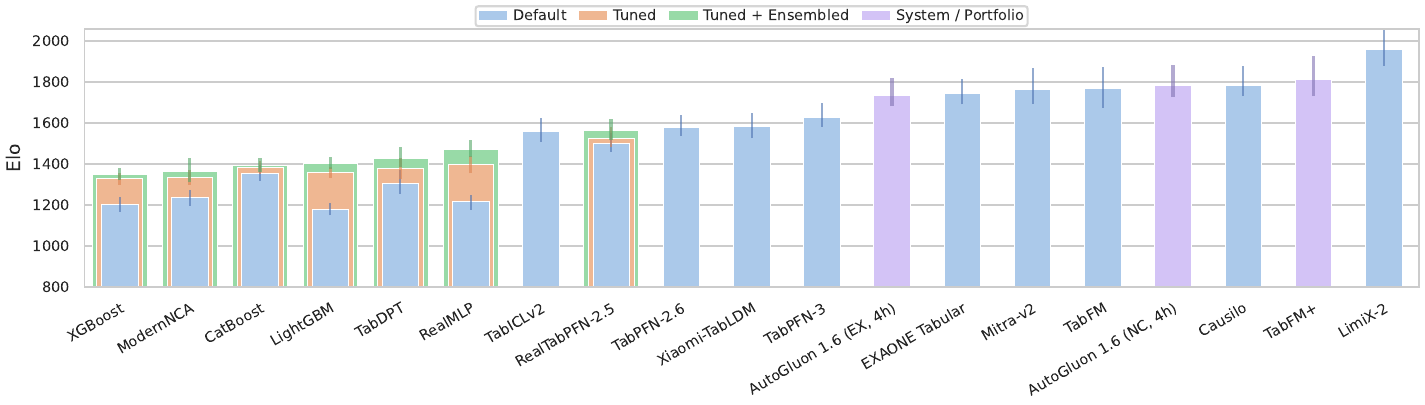}
    \caption{\textbf{Performance on the TabArena benchmark.} Baseline results are reported under the default, tuned, and tuned-plus-ensembled configurations. \ourA~with default configuration achieves an Elo score of $1963$, outperforming all compared foundation models, including TabFM, TabPFN-3, and EXAONE Tabular. \ourA~also surpasses AutoGluon under its noncommercial 4h configuration.}
    \label{fig:tabarena_leaderboard}
\end{figure}

\paragraph{Overall performance.}
\Cref{fig:tabarena_leaderboard} and~\Cref{tab:selected-overall} report performance on the full TabArena benchmark. Across all four predictive metrics reported in the table, \ourA~ranks first among the compared methods. It attains the highest Elo score of 1963, exceeding the Elo runner-up, TabFM+, by 146.8 points before rounding, and yields an improvability of 3.6\% versus 6.4\% for TabFM+. Consistently, \ourA~achieves an average rank of 5.0, compared with 9.1 for TabFM+, and records an aggregated win count of 21.7 against 5.1 for TabFM+, corresponding to approximately 4.3 times as many aggregated wins. 

Win rate averages pairwise comparisons against all other methods in the corresponding table, first across evaluation splits within each dataset and then equally across datasets, with ties counted as half a win. The aggregated win count instead assigns one unit of credit per split to the best-performing method, shares it equally among tied winners, averages within each dataset, and sums across datasets. Elo point estimates are rounded to integers in the tables. Elo differences and relative comparisons are computed from the unrounded results.

\paragraph{Classification performance.}
\Cref{tab:selected-class} reports results on the 38 classification datasets. \ourA~ranks first across all four predictive metrics reported in the table. It attains an Elo of 1928, an improvability of 4.7\%, an average rank of 5.7, an aggregated win count of 12.8, and an average pairwise win rate of 94.8\%. These results indicate strong aggregate performance across the classification tasks.

\paragraph{Regression performance.}
\Cref{tab:selected-regr} reports results on the 13 regression datasets. \ourA~again ranks first across all four predictive metrics reported in the table, achieving an Elo of 2297, an improvability of 0.3\%, an average rank of 2.7, an aggregated win count of 8.9, and an average pairwise win rate of 98.1\%.  TabFM+ is the Elo runner-up at 2064, followed by AutoGluon 1.6 (noncommercial, 4h) at 2060. These results indicate strong performance on the regression subset.

\begingroup
\fontsize{7.6}{9.2}\selectfont
\setlength{\tabcolsep}{1.65pt}
\renewcommand{\arraystretch}{1.10}
\setlength{\LTpre}{8pt}
\setlength{\LTpost}{8pt}
\setlength{\LTcapwidth}{\linewidth}
\setlength{\LTleft}{0pt plus 1fil}
\setlength{\LTright}{0pt plus 1fil}
\begin{longtable}{@{\extracolsep{\fill}}llccc@{\hspace{\dimexpr0.6em-0.2pt\relax}{\color{black}\vrule width 0.4pt}\hspace{\dimexpr0.6em-0.2pt\relax}}llccc@{}}
\caption{\textbf{Performance on the TabArena benchmark.} Models are ranked by Elo, improvability, average rank, and aggregated win count are also reported. \ourA~ranks first on all four metrics, with an Elo of 1963 (146.8 points above TabFM+ before rounding). Compared with TabFM+, its improvability is 3.6\% versus 6.4\%, its average rank is 5.0 versus 9.1, and its aggregated win count is 21.7 versus 5.1.}
\label{tab:selected-overall}\\[4pt]
\toprule
\textbf{Model} & \makecell{\textbf{Elo}\\$\uparrow$} & \makecell{\textbf{Improv-}\\\textbf{ability} $\downarrow$} & \makecell{\textbf{Avg. rank}\\$\downarrow$} & \makecell{\textbf{\#wins}\\$\uparrow$} & \textbf{Model} & \makecell{\textbf{Elo}\\$\uparrow$} & \makecell{\textbf{Improv-}\\\textbf{ability} $\downarrow$} & \makecell{\textbf{Avg. rank}\\$\downarrow$} & \makecell{\textbf{\#wins}\\$\uparrow$} \\
\midrule
\endfirsthead
\multicolumn{10}{@{}l}{\textbf{Table~\ref{tab:selected-overall} (continued)}}\\[3pt]
\toprule
\textbf{Model} & \makecell{\textbf{Elo}\\$\uparrow$} & \makecell{\textbf{Improv-}\\\textbf{ability} $\downarrow$} & \makecell{\textbf{Avg. rank}\\$\downarrow$} & \makecell{\textbf{\#wins}\\$\uparrow$} & \textbf{Model} & \makecell{\textbf{Elo}\\$\uparrow$} & \makecell{\textbf{Improv-}\\\textbf{ability} $\downarrow$} & \makecell{\textbf{Avg. rank}\\$\downarrow$} & \makecell{\textbf{\#wins}\\$\uparrow$} \\
\midrule
\endhead
\midrule\multicolumn{10}{r@{}}{\scriptsize Continued on next page}\\
\endfoot
\bottomrule
\endlastfoot

\cellcolor{evalrow}LimiX-2 (D) & \cellcolor{evalrow}\textcolor{gold}{\textbf{1963$_{-83,+137}$}} & \cellcolor{evalrow}\textcolor{gold}{\textbf{3.6\%}} & \cellcolor{evalrow}\textcolor{gold}{\textbf{5.0}} & \cellcolor{evalrow}\textcolor{gold}{\textbf{21.7}} & TabM (D) & 1282$_{-42,+43}$ & 19.1\% & 47.7 & 0.0 \\
TabFM+ & \textcolor{silver}{{1816$_{-84,+111}$}} & \textcolor{silver}{{6.4\%}} & \textcolor{silver}{{9.1}} & \textcolor{silver}{{5.1}} & iLTM (T) & 1280$_{-32,+35}$ & 19.2\% & 47.9 & 0.1 \\
Causilo (D) & \textcolor{bronze}{{1788$_{-56,+91}$}} & 9.1\% & \textcolor{bronze}{{10.1}} & 1.3 & BetaTabPFN (D) & 1272$_{-57,+58}$ & 20.5\% & 48.8 & 0.0 \\
AutoGluon 1.6 (NC, 4h) & 1787$_{-59,+101}$ & 8.8\% & 10.2 & 0.9 & TabPFNv2 (T) & 1272$_{-57,+58}$ & 20.2\% & 48.8 & 0.1 \\
TabFM (D) & 1772$_{-96,+102}$ & \textcolor{bronze}{{6.7\%}} & 10.7 & \textcolor{bronze}{{5.0}} & SAP-RPT-OSS (D) & 1268$_{-58,+57}$ & 20.4\% & 49.1 & 0.6 \\
Mitra-v2 (D) & 1768$_{-74,+104}$ & 8.5\% & 10.9 & 3.0 & TorchMLP (T+E) & 1268$_{-45,+46}$ & 19.2\% & 49.1 & 0.0 \\
EXAONE Tabular (D) & 1746$_{-54,+71}$ & 9.7\% & 11.9 & 1.5 & EBM (T+E) & 1253$_{-37,+37}$ & 20.7\% & 50.7 & 0.0 \\
AutoGluon 1.6 (EX, 4h) & 1737$_{-53,+85}$ & 9.4\% & 12.3 & 0.8 & TabPFNv2 (D) & 1245$_{-63,+66}$ & 20.9\% & 51.5 & 0.3 \\
AutoGluon 1.5 (EX, 4h) & 1647$_{-59,+68}$ & 10.4\% & 16.9 & 1.0 & ModernNCA (D) & 1238$_{-40,+38}$ & 21.2\% & 52.2 & 0.3 \\
TabPFN-3 (D) & 1631$_{-49,+69}$ & 11.8\% & 17.9 & 0.4 & EBM (T) & 1220$_{-41,+45}$ & 21.3\% & 54.0 & 0.0 \\
Xiaomi-TabLDM (D) & 1586$_{-59,+65}$ & 12.3\% & 20.8 & 0.6 & RealMLP (D) & 1217$_{-38,+32}$ & 20.6\% & 54.3 & 0.1 \\
TabPFN-2.6 (D) & 1579$_{-42,+59}$ & 13.1\% & 21.2 & 0.1 & XGBoost (D) & 1205$_{-36,+34}$ & 20.8\% & 55.6 & 0.0 \\
RealTabPFN-2.5 (T+E) & 1564$_{-47,+59}$ & 12.8\% & 22.3 & 0.2 & ExtraTrees (T+E) & 1201$_{-44,+46}$ & 21.7\% & 55.9 & 0.0 \\
TabICLv2 (D) & 1563$_{-54,+62}$ & 12.6\% & 22.4 & 0.1 & TorchMLP (T) & 1200$_{-42,+41}$ & 20.9\% & 56.1 & 0.0 \\
RealTabPFN-2.5 (T) & 1528$_{-44,+54}$ & 13.5\% & 25.0 & 0.1 & FastaiMLP (T+E) & 1192$_{-59,+54}$ & 21.6\% & 56.8 & 0.0 \\
RealTabPFN-2.5 (D) & 1501$_{-40,+51}$ & 14.0\% & 27.1 & 0.1 & EBM (D) & 1192$_{-48,+46}$ & 22.2\% & 56.8 & 0.1 \\
AutoGluon 1.4 (best, 4h) & 1479$_{-46,+44}$ & 15.0\% & 28.9 & 0.0 & APLR (T+E) & 1186$_{-56,+41}$ & 22.3\% & 57.4 & 0.0 \\
RealMLP (T+E) & 1474$_{-41,+45}$ & 15.1\% & 29.3 & 0.1 & LightGBM (D) & 1180$_{-30,+32}$ & 21.3\% & 58.0 & 0.0 \\
TabDPT-Turbo (D) & 1432$_{-47,+49}$ & 16.2\% & 33.0 & 0.2 & ExtraTrees (T) & 1177$_{-54,+43}$ & 22.5\% & 58.3 & 0.2 \\
TabDPT (T+E) & 1431$_{-45,+56}$ & 16.0\% & 33.1 & 0.4 & CTBoost (D) & 1171$_{-65,+60}$ & 23.2\% & 58.9 & 0.1 \\
TabM (T+E) & 1421$_{-37,+45}$ & 16.3\% & 34.0 & 0.0 & RandomForest (T+E) & 1169$_{-44,+52}$ & 22.7\% & 59.1 & 0.1 \\
LightGBM (T+E) & 1405$_{-27,+30}$ & 17.2\% & 35.5 & 0.0 & Nori-30M (D) & 1156$_{-82,+73}$ & 25.0\% & 60.4 & 0.1 \\
RealMLP (T) & 1402$_{-47,+39}$ & 16.4\% & 35.8 & 0.1 & APLR (T) & 1145$_{-56,+46}$ & 23.4\% & 61.4 & 0.0 \\
CatBoost (T+E) & 1395$_{-32,+36}$ & 16.8\% & 36.4 & 0.1 & Nori (D) & 1145$_{-77,+70}$ & 25.2\% & 61.4 & 0.0 \\
CatBoost (T) & 1384$_{-32,+33}$ & 17.0\% & 37.5 & 0.0 & FastaiMLP (T) & 1137$_{-58,+53}$ & 23.0\% & 62.2 & 0.1 \\
TabDPT (T) & 1380$_{-52,+53}$ & 17.1\% & 37.9 & 0.1 & RandomForest (T) & 1134$_{-46,+47}$ & 23.4\% & 62.5 & 0.2 \\
iLTM (T+E) & 1379$_{-38,+38}$ & 17.4\% & 38.0 & 0.0 & TabSTAR (T) & 1088$_{-83,+75}$ & 27.3\% & 66.7 & 0.8 \\
TabM (T) & 1371$_{-37,+43}$ & 17.1\% & 38.8 & 0.0 & iLTM (D) & 1088$_{-57,+47}$ & 24.9\% & 66.8 & 0.1 \\
ModernNCA (T+E) & 1364$_{-52,+66}$ & 17.8\% & 39.5 & 0.1 & TabSTAR (T+E) & 1087$_{-84,+79}$ & 27.3\% & 66.8 & 0.9 \\
LightGBM (T) & 1362$_{-27,+25}$ & 17.8\% & 39.6 & 0.0 & PerpetualBooster (T+E) & 1086$_{-45,+46}$ & 28.1\% & 66.9 & 0.0 \\
ChimeraBoost (T+E) & 1360$_{-53,+44}$ & 18.1\% & 39.8 & 0.8 & OrionMSP (D) & 1086$_{-52,+54}$ & 25.5\% & 66.9 & 0.0 \\
CatBoost (D) & 1357$_{-38,+39}$ & 17.6\% & 40.2 & 0.1 & APLR (D) & 1079$_{-80,+51}$ & 26.2\% & 67.6 & 0.0 \\
XGBoost (T+E) & 1353$_{-30,+29}$ & 17.9\% & 40.5 & 0.0 & TorchMLP (D) & 1071$_{-48,+37}$ & 24.7\% & 68.2 & 0.1 \\
LimiX (D) & 1344$_{-56,+72}$ & 17.7\% & 41.4 & 0.3 & PerpetualBooster (T) & 1052$_{-46,+47}$ & 29.4\% & 69.9 & 0.0 \\
ModernNCA (T) & 1336$_{-37,+36}$ & 18.1\% & 42.2 & 0.3 & xRFM (D) & 1039$_{-69,+55}$ & 28.2\% & 70.9 & 0.0 \\
ChimeraBoost (T) & 1333$_{-51,+43}$ & 18.5\% & 42.5 & 0.3 & TabFlex (D) & 1010$_{-74,+66}$ & 29.3\% & 73.2 & 0.1 \\
TabSwift (D) & 1332$_{-49,+60}$ & 18.2\% & 42.7 & 0.0 & ExtraTrees (D) & 1008$_{-61,+48}$ & 27.9\% & 73.4 & 0.0 \\
XGBoost (T) & 1330$_{-30,+28}$ & 18.2\% & 42.8 & 0.0 & FastaiMLP (D) & 1005$_{-61,+59}$ & 27.4\% & 73.6 & 0.1 \\
xRFM (T+E) & 1330$_{-44,+44}$ & 18.4\% & 42.9 & 0.1 & RandomForest (D) & 1000$_{-47,+43}$ & 28.0\% & 73.9 & 0.0 \\
TabPFNv2 (T+E) & 1325$_{-66,+63}$ & 18.8\% & 43.3 & 0.0 & KNN (T+E) & 991$_{-81,+62}$ & 29.5\% & 74.6 & 0.2 \\
CTBoost (T+E) & 1324$_{-42,+39}$ & 18.8\% & 43.5 & 0.0 & TabSTAR (D) & 990$_{-101,+90}$ & 32.0\% & 74.7 & 0.4 \\
Mitra (D) & 1315$_{-64,+62}$ & 19.2\% & 44.4 & 0.1 & Linear (T+E) & 956$_{-102,+61}$ & 35.2\% & 77.1 & 0.0 \\
TabDPT (D) & 1308$_{-53,+62}$ & 19.3\% & 45.0 & 0.1 & PerpetualBooster (D) & 934$_{-61,+41}$ & 33.3\% & 78.5 & 0.1 \\
TabICL (D) & 1306$_{-57,+49}$ & 19.2\% & 45.3 & 0.0 & Linear (T) & 932$_{-109,+67}$ & 35.7\% & 78.6 & 0.0 \\
ChimeraBoost (D) & 1299$_{-55,+42}$ & 19.3\% & 46.0 & 0.0 & KNN (T) & 887$_{-101,+61}$ & 34.3\% & 81.3 & 0.1 \\
CTBoost (T) & 1293$_{-44,+37}$ & 19.3\% & 46.6 & 0.0 & Linear (D) & 859$_{-118,+68}$ & 38.2\% & 82.8 & 0.1 \\
xRFM (T) & 1286$_{-43,+42}$ & 19.5\% & 47.3 & 0.1 & KNN (D) & 642$_{-99,+85}$ & 47.1\% & 90.2 & 0.0 \\
\end{longtable}
\endgroup

\begingroup
\fontsize{7.6}{9.2}\selectfont
\setlength{\tabcolsep}{1.65pt}
\renewcommand{\arraystretch}{1.10}
\setlength{\LTpre}{8pt}
\setlength{\LTpost}{8pt}
\setlength{\LTcapwidth}{\linewidth}
\setlength{\LTleft}{0pt plus 1fil}
\setlength{\LTright}{0pt plus 1fil}
\begin{longtable}{@{\extracolsep{\fill}}llccc@{\hspace{\dimexpr0.6em-0.2pt\relax}{\color{black}\vrule width 0.4pt}\hspace{\dimexpr0.6em-0.2pt\relax}}llccc@{}}
\caption{\textbf{Performance on classification tasks on the TabArena benchmark.} Models are ranked by Elo, improvability, average rank, and aggregated win count are also reported. \ourA~ranks first on all four metrics, with an Elo of 1928 (132.4 points above TabFM+ before rounding). Its improvability is 4.7\%, approximately 36.4\% lower than that of TabFM+ (7.5\%). Its average rank is 5.7 versus 9.9, and its aggregated win count is 12.8 versus 4.7.}
\label{tab:selected-class}\\[4pt]
\toprule
\textbf{Model} & \makecell{\textbf{Elo}\\$\uparrow$} & \makecell{\textbf{Improv-}\\\textbf{ability} $\downarrow$} & \makecell{\textbf{Avg. rank}\\$\downarrow$} & \makecell{\textbf{\#wins}\\$\uparrow$} & \textbf{Model} & \makecell{\textbf{Elo}\\$\uparrow$} & \makecell{\textbf{Improv-}\\\textbf{ability} $\downarrow$} & \makecell{\textbf{Avg. rank}\\$\downarrow$} & \makecell{\textbf{\#wins}\\$\uparrow$} \\
\midrule
\endfirsthead
\multicolumn{10}{@{}l}{\textbf{Table~\ref{tab:selected-class} (continued)}}\\[3pt]
\toprule
\textbf{Model} & \makecell{\textbf{Elo}\\$\uparrow$} & \makecell{\textbf{Improv-}\\\textbf{ability} $\downarrow$} & \makecell{\textbf{Avg. rank}\\$\downarrow$} & \makecell{\textbf{\#wins}\\$\uparrow$} & \textbf{Model} & \makecell{\textbf{Elo}\\$\uparrow$} & \makecell{\textbf{Improv-}\\\textbf{ability} $\downarrow$} & \makecell{\textbf{Avg. rank}\\$\downarrow$} & \makecell{\textbf{\#wins}\\$\uparrow$} \\
\midrule
\endhead
\midrule\multicolumn{10}{r@{}}{\scriptsize Continued on next page}\\
\endfoot
\bottomrule
\endlastfoot

\cellcolor{evalrow}LimiX-2 (D) & \cellcolor{evalrow}\textcolor{gold}{\textbf{1928$_{-97,+164}$}} & \cellcolor{evalrow}\textcolor{gold}{\textbf{4.7\%}} & \cellcolor{evalrow}\textcolor{gold}{\textbf{5.7}} & \cellcolor{evalrow}\textcolor{gold}{\textbf{12.8}} & TabPFNv2 (T) & 1304$_{-77,+70}$ & 22.9\% & 46.7 & 0.1 \\
TabFM+ & \textcolor{silver}{{1795$_{-86,+118}$}} & \textcolor{silver}{{7.5\%}} & \textcolor{silver}{{9.9}} & \textcolor{bronze}{{4.7}} & TabM (D) & 1304$_{-43,+53}$ & 21.5\% & 46.8 & 0.0 \\
Causilo (D) & \textcolor{bronze}{{1767$_{-71,+90}$}} & 11.0\% & \textcolor{bronze}{{11.1}} & 1.0 & EBM (T+E) & 1301$_{-37,+39}$ & 22.0\% & 47.1 & 0.0 \\
AutoGluon 1.6 (NC, 4h) & 1760$_{-54,+87}$ & 10.6\% & 11.4 & 0.8 & TorchMLP (T+E) & 1297$_{-39,+42}$ & 21.2\% & 47.5 & 0.0 \\
TabFM (D) & 1759$_{-113,+123}$ & \textcolor{bronze}{{7.5\%}} & 11.4 & \textcolor{silver}{{4.8}} & SAP-RPT-OSS (D) & 1291$_{-70,+53}$ & 22.8\% & 48.1 & 0.5 \\
Mitra-v2 (D) & 1754$_{-85,+125}$ & 10.2\% & 11.7 & 2.8 & iLTM (T) & 1285$_{-37,+42}$ & 21.8\% & 48.7 & 0.1 \\
EXAONE Tabular (D) & 1752$_{-58,+82}$ & 11.2\% & 11.7 & 1.5 & TabPFNv2 (D) & 1281$_{-81,+71}$ & 23.4\% & 49.1 & 0.3 \\
AutoGluon 1.6 (EX, 4h) & 1710$_{-43,+62}$ & 11.2\% & 13.8 & 0.6 & xRFM (T) & 1280$_{-61,+55}$ & 22.5\% & 49.2 & 0.1 \\
AutoGluon 1.5 (EX, 4h) & 1652$_{-70,+81}$ & 11.7\% & 17.0 & 0.8 & EBM (T) & 1268$_{-39,+44}$ & 22.7\% & 50.4 & 0.0 \\
TabPFN-3 (D) & 1627$_{-65,+76}$ & 14.0\% & 18.6 & 0.4 & TabDPT (D) & 1263$_{-58,+63}$ & 23.1\% & 50.9 & 0.1 \\
Xiaomi-TabLDM (D) & 1577$_{-55,+70}$ & 14.6\% & 22.0 & 0.5 & FastaiMLP (T+E) & 1250$_{-66,+60}$ & 23.0\% & 52.2 & 0.0 \\
TabPFN-2.6 (D) & 1576$_{-55,+60}$ & 15.3\% & 22.1 & 0.1 & EBM (D) & 1246$_{-49,+42}$ & 23.6\% & 52.7 & 0.1 \\
TabICLv2 (D) & 1569$_{-63,+70}$ & 14.6\% & 22.6 & 0.1 & ModernNCA (D) & 1244$_{-42,+42}$ & 24.0\% & 52.8 & 0.3 \\
RealTabPFN-2.5 (T+E) & 1557$_{-66,+65}$ & 15.1\% & 23.5 & 0.1 & APLR (T+E) & 1239$_{-58,+45}$ & 23.7\% & 53.3 & 0.0 \\
RealTabPFN-2.5 (T) & 1537$_{-58,+67}$ & 15.7\% & 25.0 & 0.1 & CTBoost (D) & 1233$_{-65,+60}$ & 24.1\% & 53.9 & 0.1 \\
RealTabPFN-2.5 (D) & 1526$_{-52,+52}$ & 16.0\% & 25.9 & 0.1 & RealMLP (D) & 1232$_{-37,+39}$ & 23.2\% & 54.1 & 0.1 \\
AutoGluon 1.4 (best, 4h) & 1481$_{-50,+64}$ & 17.1\% & 29.6 & 0.0 & XGBoost (D) & 1222$_{-45,+41}$ & 23.0\% & 55.0 & 0.0 \\
RealMLP (T+E) & 1463$_{-37,+52}$ & 17.7\% & 31.2 & 0.1 & TorchMLP (T) & 1220$_{-38,+46}$ & 23.2\% & 55.2 & 0.0 \\
TabM (T+E) & 1442$_{-42,+53}$ & 18.6\% & 33.1 & 0.0 & ExtraTrees (T+E) & 1206$_{-53,+56}$ & 23.9\% & 56.6 & 0.0 \\
TabDPT-Turbo (D) & 1419$_{-56,+72}$ & 19.0\% & 35.4 & 0.1 & APLR (T) & 1195$_{-61,+49}$ & 24.9\% & 57.6 & 0.0 \\
LightGBM (T+E) & 1412$_{-29,+46}$ & 19.3\% & 36.0 & 0.0 & FastaiMLP (T) & 1188$_{-68,+56}$ & 24.7\% & 58.3 & 0.1 \\
TabICL (D) & 1406$_{-53,+49}$ & 19.5\% & 36.6 & 0.0 & LightGBM (D) & 1184$_{-42,+49}$ & 23.9\% & 58.7 & 0.0 \\
CatBoost (T+E) & 1397$_{-43,+51}$ & 19.0\% & 37.4 & 0.1 & ExtraTrees (T) & 1180$_{-62,+61}$ & 24.9\% & 59.1 & 0.2 \\
LimiX (D) & 1396$_{-73,+83}$ & 19.3\% & 37.6 & 0.3 & RandomForest (T+E) & 1180$_{-66,+63}$ & 25.0\% & 59.2 & 0.1 \\
TabM (T) & 1395$_{-42,+48}$ & 19.4\% & 37.6 & 0.0 & iLTM (D) & 1147$_{-56,+57}$ & 26.2\% & 62.2 & 0.1 \\
TabDPT (T+E) & 1393$_{-48,+70}$ & 19.2\% & 37.8 & 0.1 & RandomForest (T) & 1146$_{-68,+68}$ & 25.7\% & 62.3 & 0.2 \\
RealMLP (T) & 1393$_{-43,+50}$ & 19.0\% & 37.8 & 0.1 & APLR (D) & 1141$_{-67,+59}$ & 27.3\% & 62.7 & 0.0 \\
iLTM (T+E) & 1388$_{-40,+44}$ & 19.4\% & 38.3 & 0.0 & TabSTAR (T) & 1135$_{-96,+75}$ & 27.9\% & 63.3 & 0.7 \\
CatBoost (T) & 1388$_{-40,+48}$ & 19.2\% & 38.3 & 0.0 & TabSTAR (T+E) & 1131$_{-94,+74}$ & 28.0\% & 63.6 & 0.7 \\
CatBoost (D) & 1374$_{-44,+45}$ & 19.6\% & 39.7 & 0.1 & OrionMSP (D) & 1117$_{-66,+62}$ & 28.0\% & 64.9 & 0.0 \\
LightGBM (T) & 1372$_{-31,+37}$ & 20.0\% & 39.9 & 0.0 & TorchMLP (D) & 1087$_{-54,+45}$ & 27.2\% & 67.4 & 0.1 \\
CTBoost (T+E) & 1371$_{-55,+51}$ & 20.2\% & 40.0 & 0.0 & PerpetualBooster (T+E) & 1066$_{-61,+56}$ & 32.5\% & 69.2 & 0.0 \\
ChimeraBoost (T+E) & 1371$_{-55,+52}$ & 20.3\% & 40.0 & 0.0 & FastaiMLP (D) & 1044$_{-73,+70}$ & 29.3\% & 70.9 & 0.1 \\
XGBoost (T+E) & 1362$_{-45,+44}$ & 20.1\% & 40.9 & 0.0 & Linear (T+E) & 1038$_{-88,+88}$ & 34.0\% & 71.4 & 0.0 \\
BetaTabPFN (D) & 1360$_{-48,+47}$ & 21.4\% & 41.0 & 0.0 & TabSTAR (D) & 1034$_{-118,+85}$ & 33.3\% & 71.7 & 0.3 \\
Mitra (D) & 1352$_{-80,+76}$ & 21.5\% & 41.9 & 0.1 & PerpetualBooster (T) & 1032$_{-58,+55}$ & 33.9\% & 71.8 & 0.0 \\
TabPFNv2 (T+E) & 1352$_{-73,+85}$ & 21.3\% & 41.9 & 0.0 & KNN (T+E) & 1025$_{-89,+74}$ & 31.7\% & 72.4 & 0.2 \\
ModernNCA (T) & 1350$_{-38,+53}$ & 20.3\% & 42.1 & 0.3 & TabFlex (D) & 1014$_{-102,+82}$ & 33.2\% & 73.1 & 0.1 \\
ModernNCA (T+E) & 1347$_{-69,+73}$ & 20.5\% & 42.4 & 0.0 & Linear (T) & 1012$_{-94,+88}$ & 34.6\% & 73.3 & 0.0 \\
ChimeraBoost (T) & 1344$_{-53,+52}$ & 20.7\% & 42.7 & 0.0 & xRFM (D) & 1009$_{-86,+68}$ & 32.4\% & 73.5 & 0.0 \\
CTBoost (T) & 1341$_{-52,+47}$ & 20.7\% & 43.0 & 0.0 & RandomForest (D) & 1000$_{-65,+59}$ & 31.4\% & 74.1 & 0.0 \\
TabDPT (T) & 1339$_{-51,+60}$ & 20.5\% & 43.2 & 0.1 & ExtraTrees (D) & 992$_{-71,+73}$ & 31.4\% & 74.6 & 0.0 \\
TabSwift (D) & 1338$_{-57,+55}$ & 21.1\% & 43.3 & 0.0 & Linear (D) & 937$_{-107,+84}$ & 37.2\% & 78.1 & 0.1 \\
XGBoost (T) & 1337$_{-38,+38}$ & 20.5\% & 43.4 & 0.0 & PerpetualBooster (D) & 928$_{-79,+59}$ & 37.9\% & 78.6 & 0.1 \\
xRFM (T+E) & 1316$_{-47,+51}$ & 21.3\% & 45.5 & 0.1 & KNN (T) & 913$_{-116,+85}$ & 37.4\% & 79.5 & 0.1 \\
ChimeraBoost (D) & 1315$_{-66,+53}$ & 21.6\% & 45.6 & 0.0 & KNN (D) & 580$_{-118,+108}$ & 52.3\% & 89.7 & 0.0 \\
\end{longtable}
\endgroup

\begingroup
\fontsize{7.6}{9.2}\selectfont
\setlength{\tabcolsep}{1.65pt}
\renewcommand{\arraystretch}{1.10}
\setlength{\LTpre}{8pt}
\setlength{\LTpost}{8pt}
\setlength{\LTcapwidth}{\linewidth}
\setlength{\LTleft}{0pt plus 1fil}
\setlength{\LTright}{0pt plus 1fil}
\begin{longtable}{@{\extracolsep{\fill}}llccc@{\hspace{\dimexpr0.6em-0.2pt\relax}{\color{black}\vrule width 0.4pt}\hspace{\dimexpr0.6em-0.2pt\relax}}llccc@{}}
\caption{\textbf{Performance on regression tasks on the TabArena benchmark.} Models are ranked by Elo, improvability, average rank, and aggregated win count are also reported. \ourA~ranks first on all four metrics, with an Elo of 2297, exceeding TabFM+ and AutoGluon 1.6 (noncommercial, 4h) by 233.7 and 237.6 points, respectively, before rounding. Compared with those two methods, respectively, its improvability is 0.3\% versus 3.4\% and 3.8\%, its average rank is 2.7 versus 6.5 and 6.5, and its aggregated win count is 8.9 versus 0.4 and 0.1.}
\label{tab:selected-regr}\\[4pt]
\toprule
\textbf{Model} & \makecell{\textbf{Elo}\\$\uparrow$} & \makecell{\textbf{Improv-}\\\textbf{ability} $\downarrow$} & \makecell{\textbf{Avg. rank}\\$\downarrow$} & \makecell{\textbf{\#wins}\\$\uparrow$} & \textbf{Model} & \makecell{\textbf{Elo}\\$\uparrow$} & \makecell{\textbf{Improv-}\\\textbf{ability} $\downarrow$} & \makecell{\textbf{Avg. rank}\\$\downarrow$} & \makecell{\textbf{\#wins}\\$\uparrow$} \\
\midrule
\endfirsthead
\multicolumn{10}{@{}l}{\textbf{Table~\ref{tab:selected-regr} (continued)}}\\[3pt]
\toprule
\textbf{Model} & \makecell{\textbf{Elo}\\$\uparrow$} & \makecell{\textbf{Improv-}\\\textbf{ability} $\downarrow$} & \makecell{\textbf{Avg. rank}\\$\downarrow$} & \makecell{\textbf{\#wins}\\$\uparrow$} & \textbf{Model} & \makecell{\textbf{Elo}\\$\uparrow$} & \makecell{\textbf{Improv-}\\\textbf{ability} $\downarrow$} & \makecell{\textbf{Avg. rank}\\$\downarrow$} & \makecell{\textbf{\#wins}\\$\uparrow$} \\
\midrule
\endhead
\midrule\multicolumn{10}{r@{}}{\scriptsize Continued on next page}\\
\endfoot
\bottomrule
\endlastfoot

\cellcolor{evalrow}LimiX-2 (D) & \cellcolor{evalrow}\textcolor{gold}{\textbf{2297$_{-176,+322}$}} & \cellcolor{evalrow}\textcolor{gold}{\textbf{0.3\%}} & \cellcolor{evalrow}\textcolor{gold}{\textbf{2.7}} & \cellcolor{evalrow}\textcolor{gold}{\textbf{8.9}} & TabM (D) & 1274$_{-94,+114}$ & 11.9\% & 48.7 & 0.0 \\
TabFM+ & \textcolor{silver}{{2064$_{-137,+211}$}} & \textcolor{silver}{{3.4\%}} & \textcolor{silver}{{6.5}} & \textcolor{bronze}{{0.4}} & ModernNCA (D) & 1267$_{-73,+82}$ & 13.1\% & 49.2 & 0.0 \\
AutoGluon 1.6 (NC, 4h) & \textcolor{bronze}{{2060$_{-140,+189}$}} & 3.8\% & \textcolor{bronze}{{6.5}} & 0.1 & Mitra (D) & 1263$_{-110,+126}$ & 12.7\% & 49.5 & 0.0 \\
Causilo (D) & 2028$_{-131,+233}$ & \textcolor{bronze}{{3.6\%}} & 7.3 & 0.3 & SAP-RPT-OSS (D) & 1260$_{-146,+158}$ & 13.2\% & 49.8 & 0.1 \\
AutoGluon 1.6 (EX, 4h) & 2003$_{-104,+165}$ & 4.2\% & 7.9 & 0.2 & LimiX (D) & 1247$_{-158,+177}$ & 12.8\% & 50.8 & 0.0 \\
TabFM (D) & 1978$_{-95,+172}$ & 4.4\% & 8.6 & 0.3 & TabPFNv2 (T) & 1232$_{-131,+146}$ & 12.3\% & 51.9 & 0.0 \\
Mitra-v2 (D) & 1976$_{-123,+221}$ & 3.7\% & 8.7 & 0.3 & TorchMLP (T+E) & 1229$_{-108,+102}$ & 13.4\% & 52.2 & 0.0 \\
EXAONE Tabular (D) & 1869$_{-91,+133}$ & 5.5\% & 12.2 & 0.1 & CTBoost (T+E) & 1229$_{-117,+82}$ & 14.9\% & 52.2 & 0.0 \\
TabPFN-3 (D) & 1782$_{-125,+209}$ & 5.2\% & 15.7 & 0.0 & ExtraTrees (T+E) & 1220$_{-103,+109}$ & 15.2\% & 52.9 & 0.0 \\
AutoGluon 1.5 (EX, 4h) & 1764$_{-77,+128}$ & 6.6\% & 16.5 & 0.2 & RealMLP (D) & 1218$_{-88,+100}$ & 13.0\% & 53.1 & 0.0 \\
Xiaomi-TabLDM (D) & 1759$_{-174,+237}$ & 5.6\% & 16.7 & 0.0 & LightGBM (D) & 1202$_{-40,+30}$ & 13.7\% & 54.2 & 0.0 \\
Nori-30M (D) & 1749$_{-80,+122}$ & 6.2\% & 17.2 & 0.1 & ExtraTrees (T) & 1199$_{-102,+103}$ & 15.4\% & 54.5 & 0.0 \\
TabPFN-2.6 (D) & 1719$_{-47,+91}$ & 6.7\% & 18.7 & 0.0 & CTBoost (T) & 1190$_{-120,+87}$ & 15.3\% & 55.2 & 0.0 \\
RealTabPFN-2.5 (T+E) & 1718$_{-85,+116}$ & 6.2\% & 18.7 & 0.0 & TabPFNv2 (D) & 1183$_{-145,+126}$ & 13.6\% & 55.7 & 0.0 \\
TabDPT (T+E) & 1708$_{-86,+153}$ & 6.9\% & 19.2 & 0.3 & XGBoost (D) & 1180$_{-88,+88}$ & 14.4\% & 55.9 & 0.0 \\
Nori (D) & 1667$_{-78,+125}$ & 7.2\% & 21.4 & 0.0 & TorchMLP (T) & 1177$_{-106,+87}$ & 14.1\% & 56.2 & 0.0 \\
TabICLv2 (D) & 1667$_{-137,+224}$ & 6.7\% & 21.4 & 0.0 & PerpetualBooster (T+E) & 1172$_{-95,+62}$ & 15.4\% & 56.5 & 0.0 \\
TabDPT (T) & 1656$_{-66,+125}$ & 7.3\% & 22.0 & 0.0 & RandomForest (T+E) & 1149$_{-79,+70}$ & 16.2\% & 58.3 & 0.0 \\
RealMLP (T+E) & 1633$_{-59,+104}$ & 7.7\% & 23.3 & 0.0 & xRFM (D) & 1144$_{-112,+114}$ & 16.0\% & 58.7 & 0.0 \\
RealTabPFN-2.5 (T) & 1616$_{-98,+134}$ & 6.9\% & 24.3 & 0.1 & EBM (T+E) & 1143$_{-170,+127}$ & 16.6\% & 58.7 & 0.0 \\
TabDPT-Turbo (D) & 1593$_{-92,+188}$ & 7.8\% & 25.7 & 0.1 & PerpetualBooster (T) & 1120$_{-95,+52}$ & 16.4\% & 60.4 & 0.0 \\
AutoGluon 1.4 (best, 4h) & 1580$_{-80,+95}$ & 9.0\% & 26.5 & 0.0 & RandomForest (T) & 1113$_{-90,+71}$ & 16.6\% & 60.9 & 0.0 \\
TabDPT (D) & 1574$_{-66,+140}$ & 8.2\% & 26.9 & 0.0 & EBM (T) & 1102$_{-166,+125}$ & 17.2\% & 61.7 & 0.0 \\
RealMLP (T) & 1537$_{-65,+96}$ & 8.6\% & 29.3 & 0.0 & ExtraTrees (D) & 1055$_{-118,+88}$ & 17.8\% & 64.9 & 0.0 \\
RealTabPFN-2.5 (D) & 1529$_{-98,+125}$ & 8.2\% & 29.8 & 0.0 & EBM (D) & 1043$_{-162,+115}$ & 18.0\% & 65.7 & 0.0 \\
ModernNCA (T+E) & 1525$_{-117,+126}$ & 9.8\% & 30.1 & 0.0 & APLR (T+E) & 1029$_{-115,+98}$ & 18.1\% & 66.6 & 0.0 \\
CatBoost (T+E) & 1480$_{-59,+90}$ & 10.5\% & 33.2 & 0.0 & TorchMLP (D) & 1027$_{-132,+80}$ & 17.4\% & 66.8 & 0.0 \\
LightGBM (T+E) & 1473$_{-72,+77}$ & 10.9\% & 33.7 & 0.0 & FastaiMLP (T+E) & 1024$_{-116,+100}$ & 17.6\% & 67.0 & 0.0 \\
xRFM (T+E) & 1457$_{-102,+105}$ & 10.0\% & 34.8 & 0.0 & RandomForest (D) & 1000$_{-78,+46}$ & 18.1\% & 68.5 & 0.0 \\
CatBoost (T) & 1457$_{-62,+92}$ & 10.6\% & 34.8 & 0.0 & APLR (T) & 988$_{-136,+99}$ & 18.9\% & 69.2 & 0.0 \\
TabM (T+E) & 1441$_{-78,+111}$ & 9.5\% & 35.9 & 0.0 & FastaiMLP (T) & 980$_{-117,+103}$ & 18.0\% & 69.7 & 0.0 \\
iLTM (T+E) & 1429$_{-52,+66}$ & 11.8\% & 36.8 & 0.0 & CTBoost (D) & 974$_{-157,+103}$ & 20.5\% & 70.0 & 0.0 \\
LightGBM (T) & 1410$_{-73,+79}$ & 11.4\% & 38.2 & 0.0 & TabSTAR (T+E) & 955$_{-308,+228}$ & 25.4\% & 71.2 & 0.1 \\
ChimeraBoost (T+E) & 1408$_{-108,+109}$ & 11.7\% & 38.4 & \textcolor{silver}{{0.8}} & TabSTAR (T) & 942$_{-314,+229}$ & 25.5\% & 71.9 & 0.1 \\
XGBoost (T+E) & 1401$_{-40,+54}$ & 11.4\% & 38.9 & 0.0 & PerpetualBooster (D) & 938$_{-120,+84}$ & 19.8\% & 72.1 & 0.0 \\
TabSwift (D) & 1392$_{-109,+144}$ & 9.9\% & 39.6 & 0.0 & KNN (T+E) & 888$_{-172,+147}$ & 22.9\% & 74.7 & 0.0 \\
XGBoost (T) & 1380$_{-48,+63}$ & 11.5\% & 40.5 & 0.0 & iLTM (D) & 873$_{-109,+68}$ & 21.0\% & 75.5 & 0.0 \\
CatBoost (D) & 1376$_{-86,+86}$ & 12.0\% & 40.8 & 0.0 & FastaiMLP (D) & 867$_{-156,+103}$ & 22.0\% & 75.8 & 0.0 \\
xRFM (T) & 1375$_{-86,+79}$ & 10.7\% & 40.9 & 0.0 & APLR (D) & 845$_{-210,+118}$ & 23.2\% & 76.8 & 0.0 \\
ChimeraBoost (T) & 1374$_{-107,+104}$ & 12.0\% & 41.0 & 0.2 & TabSTAR (D) & 833$_{-361,+254}$ & 28.4\% & 77.3 & 0.1 \\
TabM (T) & 1369$_{-89,+102}$ & 10.4\% & 41.3 & 0.0 & KNN (T) & 797$_{-186,+142}$ & 25.2\% & 78.8 & 0.0 \\
ModernNCA (T) & 1364$_{-89,+102}$ & 11.7\% & 41.7 & 0.0 & KNN (D) & 695$_{-248,+172}$ & 32.0\% & 82.2 & 0.0 \\
iLTM (T) & 1325$_{-59,+72}$ & 11.7\% & 44.7 & 0.0 & Linear (T+E) & 479$_{-379,+128}$ & 38.8\% & 86.7 & 0.0 \\
ChimeraBoost (D) & 1316$_{-108,+112}$ & 12.8\% & 45.4 & 0.0 & Linear (T) & 446$_{-479,+149}$ & 38.9\% & 87.2 & 0.0 \\
TabPFNv2 (T+E) & 1315$_{-119,+162}$ & 11.2\% & 45.5 & 0.0 & Linear (D) & 271$_{-439,+152}$ & 41.2\% & 88.9 & 0.0 \\
\end{longtable}
\endgroup

\begin{figure}[H]
    \centering
    \includegraphics[width=0.68\linewidth]{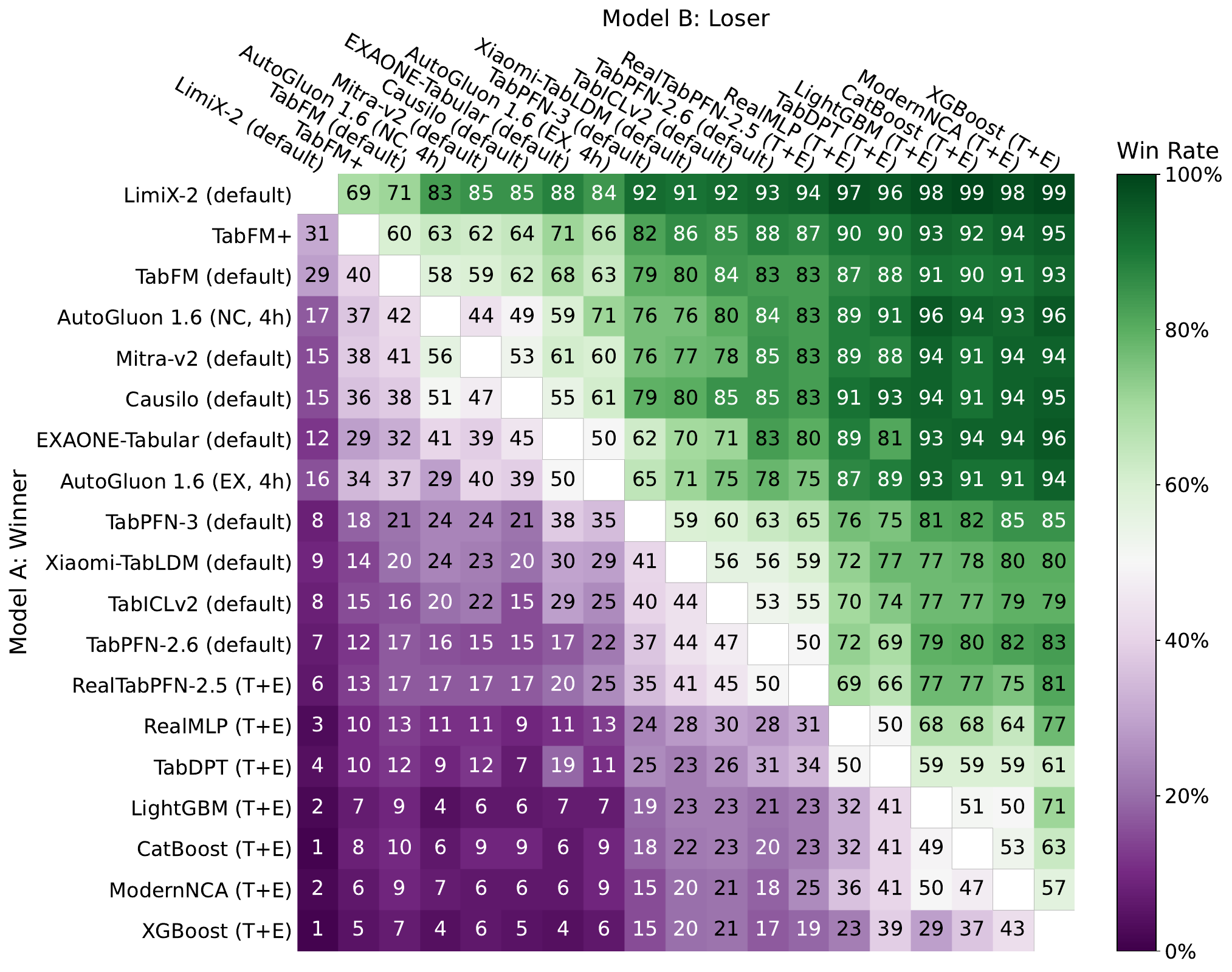}
    \caption{\textbf{Pairwise win rates on the TabArena benchmark.} Entry $(i,j)$ is the pairwise win rate of method $i$ against method $j$, computed within each dataset and then averaged equally across datasets, with ties counted as half a win. All off-diagonal entries in the row corresponding to \ourA~exceed 60\%.}
    \label{fig:tabarena_winrate_matrix}
\end{figure}

\paragraph{Pairwise comparisons.}
\Cref{fig:tabarena_winrate_matrix} reports pairwise win rates for the displayed methods and configurations. Each entry $(i,j)$ is computed by first calculating the split-level win rate of method $i$ against method $j$ within each dataset, with ties counted as half a win, and then averaging equally across datasets. All off-diagonal entries in the row corresponding to \ourA~exceed 60\%, indicating that \ourA~holds a majority win rate against every displayed competitor. In particular, its win rates against TabFM+, TabFM, AutoGluon 1.6 (noncommercial, 4h), and EXAONE Tabular are 68.6\%, 71.4\%, 83.3\%, and 88.1\%, respectively. It also achieves 84.6\% against Causilo~\citep{causilo} and 84.6\% against Mitra-v2. Its win rates exceed 90\% against the remaining displayed tabular foundation models, and reach at least 95\% against the displayed tuned-and-ensembled tree-based and neural-network baselines.

\subsection{Results on TALENT}

\paragraph{Overall performance.}
\Cref{tab:talent_results} reports performance on the TALENT benchmark. 
\ourA~achieves the highest overall Elo, together with the lowest overall improvability and the largest aggregated win count among the compared methods. 
Its overall Elo reaches 1510, exceeding TabFM and AutoGluon 1.6 (EX, 4h) by 39 and 72 points, respectively. 
The improvability decreases to 6.75\%, compared with 9.39\% for TabFM, corresponding to a 28.1\% relative reduction. 
\ourA~also records an aggregated win count of 89.5, versus 46.6 for TabFM, approximately 1.9$\times$ as many wins. 
Consistent with these aggregate results, \Cref{fig:talent_rank_by_task} shows that \ourA~attains the lowest average rank in binary classification and regression, and remains competitive on multiclass classification. 
Together, these results demonstrate that its strong relative performance extends beyond TabArena to the TALENT benchmark.

\begin{table}[t]
    \centering
    \caption{\textbf{Performance on the TALENT benchmark.}
    Models are ranked by overall Elo score. We also report Elo scores for classification and regression tasks, with classification further split into binary and multiclass settings. Red boldface highlights the best result in each metric. \ourA~achieves the highest overall Elo of 1510 and exceeds the runner-up, TabFM, by 39 points. Its classification and regression Elo scores reach 1466 and 1621, surpassing TabFM by 18 and 90 points, respectively, before rounding. \ourA~also leads in binary classification with an Elo of 1444, while TabFM attains the highest multiclass Elo of 1518.
    }
    \label{tab:talent_results}
    \begin{adjustbox}{max width=\linewidth}
    \begin{tabular}{@{}llllllll@{}}
    \toprule
    \multicolumn{1}{c}{\multirow{2}{*}[-0.73em]{\textbf{Model}}}
    & \multicolumn{5}{c}{\textbf{Elo ($\uparrow$)}}
    & \makecell{\textbf{Improv-}\\\textbf{ability}}
    & \textbf{\#wins} \\
    \cmidrule(lr){2-6}
    & \textbf{Overall}
    & \makecell{\textbf{Classifi-}\\\textbf{cation}}
    & \textbf{Regression}
    & \textbf{Binary}
    & \textbf{Multiclass}
    & \textbf{($\downarrow$)}
    & \textbf{($\uparrow$)} \\
    \midrule
    \rowcolor{evalrow}
    {LimiX-2} & \textcolor{gold}{\textbf{1510${}_{-33,+36}$}} & \textcolor{gold}{\textbf{1466${}_{-33,+37}$}} & \textcolor{gold}{\textbf{1621${}_{-73,+92}$}} & \textcolor{gold}{\textbf{1444${}_{-43,+52}$}} & \textcolor{silver}{{1517${}_{-44,+58}$}} & \textcolor{gold}{\textbf{6.75\%}} & \textcolor{gold}{\textbf{89.5}} \\
    {TabFM} & \textcolor{silver}{{1471${}_{-28,+29}$}} & \textcolor{silver}{{1449${}_{-33,+36}$}} & \textcolor{bronze}{{1531${}_{-53,+59}$}} & \textcolor{silver}{{1417${}_{-37,+46}$}} & \textcolor{gold}{\textbf{1518${}_{-64,+76}$}} & \textcolor{silver}{{9.39\%}} & \textcolor{bronze}{{46.6}} \\
    {AutoGluon 1.6 (EX, 4h)} & \textcolor{bronze}{{1438${}_{-26,+29}$}} & \textcolor{bronze}{{1380${}_{-32,+32}$}} & \textcolor{silver}{{1583${}_{-53,+65}$}} & 1339${}_{-37,+39}$ & \textcolor{bronze}{{1468${}_{-43,+53}$}} & \textcolor{bronze}{{11.51\%}} & \textcolor{silver}{{47.8}} \\
    EXAONE Tabular & 1393${}_{-21,+23}$ & 1357${}_{-25,+25}$ & 1479${}_{-35,+44}$ & \textcolor{bronze}{{1369${}_{-30,+36}$}} & 1341${}_{-36,+39}$ & 15.00\% & 12.3 \\
    TabPFN-3 & 1364${}_{-22,+25}$ & 1331${}_{-26,+27}$ & 1442${}_{-48,+55}$ & 1333${}_{-31,+35}$ & 1332${}_{-41,+49}$ & 15.89\% & 15.1 \\
    TabICLv2 & 1346${}_{-22,+23}$ & 1340${}_{-26,+28}$ & 1366${}_{-33,+39}$ & 1339${}_{-34,+39}$ & 1346${}_{-42,+52}$ & 15.74\% & 11.3 \\
    Xiaomi-TabLDM & 1341${}_{-19,+21}$ & 1303${}_{-21,+23}$ & 1433${}_{-36,+44}$ & 1306${}_{-28,+30}$ & 1302${}_{-34,+39}$ & 16.68\% & 6.7 \\
    TabDPT & 1272${}_{-25,+24}$ & 1280${}_{-28,+30}$ & 1262${}_{-42,+46}$ & 1312${}_{-34,+34}$ & 1226${}_{-51,+52}$ & 19.55\% & 11.1 \\
    Mitra-v2 & 1269${}_{-21,+23}$ & 1264${}_{-27,+30}$ & 1287${}_{-44,+46}$ & 1268${}_{-30,+34}$ & 1259${}_{-47,+51}$ & 17.72\% & 7.5 \\
    LimiX-16M & 1228${}_{-20,+20}$ & 1210${}_{-23,+23}$ & 1269${}_{-41,+43}$ & 1217${}_{-26,+28}$ & 1199${}_{-39,+40}$ & 20.35\% & 6.4 \\
    CatBoost & 1094${}_{-22,+22}$ & 1095${}_{-26,+26}$ & 1094${}_{-44,+37}$ & 1107${}_{-32,+31}$ & 1073${}_{-50,+46}$ & 26.86\% & 4.3 \\
    TabR & 1076${}_{-23,+24}$ & 1079${}_{-29,+28}$ & 1072${}_{-39,+35}$ & 1079${}_{-34,+31}$ & 1080${}_{-51,+49}$ & 26.31\% & 4.0 \\
    RealMLP & 1069${}_{-22,+21}$ & 1047${}_{-26,+26}$ & 1116${}_{-41,+38}$ & 1023${}_{-33,+29}$ & 1088${}_{-45,+43}$ & 26.70\% & 3.9 \\
    ModernNCA & 1055${}_{-26,+25}$ & 1054${}_{-28,+27}$ & 1058${}_{-51,+47}$ & 1063${}_{-39,+33}$ & 1040${}_{-49,+43}$ & 26.69\% & 5.6 \\
    FT-Transformer & 1032${}_{-25,+24}$ & 1024${}_{-30,+27}$ & 1047${}_{-44,+41}$ & 1028${}_{-39,+36}$ & 1018${}_{-54,+45}$ & 29.39\% & 4.1 \\
    RandomForest & 1000${}_{-29,+28}$ & 1000${}_{-37,+34}$ & 1000${}_{-53,+47}$ & 1000${}_{-48,+42}$ & 1000${}_{-67,+59}$ & 30.08\% & 3.8 \\
    XGBoost & 999${}_{-25,+24}$ & 1008${}_{-26,+26}$ & 980${}_{-58,+49}$ & 1000${}_{-31,+29}$ & 1022${}_{-53,+43}$ & 29.41\% & 2.2 \\
    LightGBM & 974${}_{-29,+28}$ & 978${}_{-37,+34}$ & 964${}_{-58,+48}$ & 977${}_{-44,+40}$ & 979${}_{-61,+59}$ & 29.82\% & 2.7 \\
    TabM & 954${}_{-32,+28}$ & 978${}_{-35,+32}$ & 900${}_{-61,+54}$ & 967${}_{-47,+42}$ & 996${}_{-56,+49}$ & 32.40\% & 2.9 \\
    \bottomrule
    \end{tabular}
    \end{adjustbox}
\end{table}

\paragraph{Classification performance.}
\Cref{tab:talent_results} shows that \ourA~achieves an Elo rating of 1466 on classification tasks, surpassing TabFM by 18 points before rounding. 
This advantage holds in binary classification, with an Elo score of 1444 compared with 1417 for TabFM, while TabFM attains a higher multiclass Elo of 1518 versus 1517 for \ourA. 
\Cref{fig:talent_rank_by_task} provides a complementary dataset-level comparison based on accuracy, where \ourA~achieves the lowest average rank of 4.79 on binary classification and an average rank of 3.96 on multiclass classification, close to TabFM at 3.95. 
The agreement between Elo and average-rank evaluations demonstrates strong classification performance across the two task settings.

\paragraph{Regression performance.}
On regression, \ourA~attains the highest Elo of 1621, ahead of AutoGluon 1.6 (EX, 4h) at 1583 and exceeding TabFM and EXAONE Tabular by 90 and 142 points, respectively, before rounding, as shown in \Cref{tab:talent_results}. 
\Cref{fig:talent_rank_by_task} provides a complementary comparison based on RMSE, where \ourA~again achieves the lowest average rank, at 3.48. 
The consistent advantage under both Elo and dataset-level average rank further supports the strong regression performance of \ourA. 

\begin{figure}[p]
    \centering
    \includegraphics[width=0.75\linewidth]{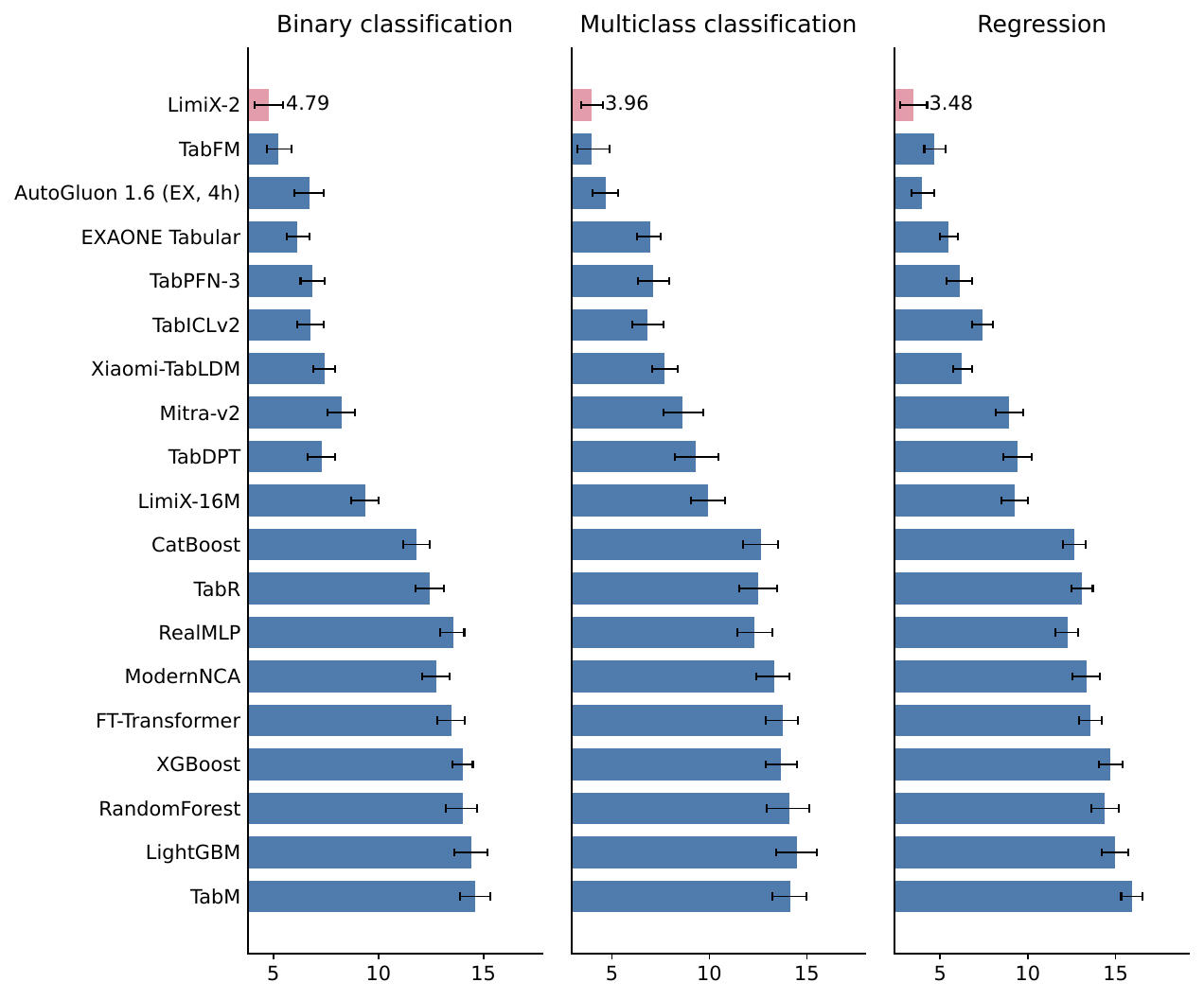}
    \caption{\textbf{Average-rank comparison on the TALENT benchmark across binary classification, multiclass classification, and regression tasks.}
    Each bar represents a method's mean rank over datasets in the corresponding task, using accuracy for classification and RMSE for regression.
    Error bars denote 95\% bootstrap CIs from 1000 dataset resamples.
    Methods are displayed in the same order across all panels.
    \ourA~is highlighted in pink, while baseline models are shown in dark blue.
    \ourA~achieves the lowest average rank in binary classification and regression, with ranks of 4.79 and 3.48, respectively, and an average rank of 3.96 on multiclass classification.
    }
    \label{fig:talent_rank_by_task}
\end{figure}

\begin{figure}[p]
    \centering
    \includegraphics[width=0.65\linewidth]{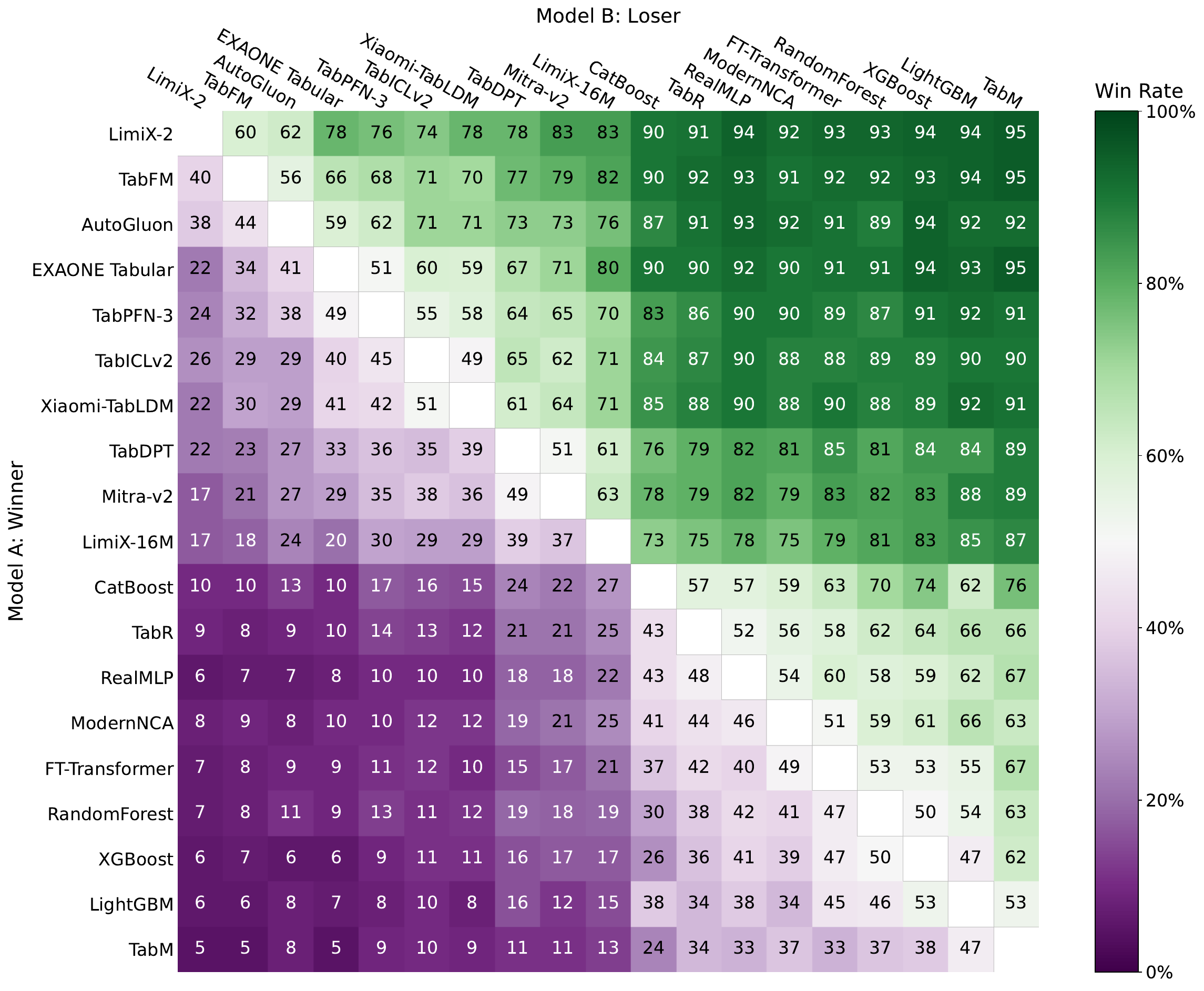}
    \caption{\textbf{Pairwise win rates on the TALENT benchmark.}
    Entry $(i,j)$ is the percentage of evaluation tasks on which method $i$ beats method $j$.
    Cells range from purple (row loses) to white (tie) to green (row wins).
    \ourA~achieves a win rate above 60\% against every competing method, ranging from 60\% against TabFM to 95\% against TabM.
    }
    \label{fig:talent_winrate_matrix}
\end{figure}

\paragraph{Pairwise comparisons.}
\Cref{fig:talent_winrate_matrix} reports pairwise win rates across evaluation tasks. 
Each entry $(i,j)$ denotes the percentage of tasks on which method $i$ outperforms method $j$. 
\ourA~wins 60\% of tasks against TabFM, 62\% against AutoGluon 1.6 (EX, 4h), and 78\% against EXAONE Tabular. 
Against the remaining tabular foundation models, its win rates range from 74\% to 83\%, while they reach 90--95\% against the remaining baselines. 
Overall, \ourA~achieves a win rate above 60\% against every competing method, showing that its leading aggregate performance is supported by broad pairwise advantages rather than gains concentrated on a small subset of datasets.

\paragraph{Meta-feature subgroups.}
\Cref{fig:talent_rank_by_metafeature} examines how model rankings vary with dataset size and dimensionality. 
\ourA~maintains among the lowest fitted ranks across most sample-size and feature-count ranges, with particularly strong performance on medium-to-large datasets. 
Across feature counts, its fitted rank also remains consistently competitive and generally improves as dimensionality increases. 
These trends indicate that the aggregate advantage of \ourA~is preserved across datasets with substantially different numbers of samples and features, rather than being driven by a narrow dataset regime.

\subsection{Results on BCCO}

\paragraph{Overall performance.}
\Cref{tab:bcco_results} reports performance on the BCCO benchmark. \ourA~achieves the highest overall Elo, along with the lowest overall improvability and the largest aggregated win count. Its overall Elo reaches 1439, exceeding AutoGluon 1.6 (EX, 4h), TabFM, and LimiX-16M by 64, 71, and 209 points, respectively. The improvability decreases to 6.79\%, compared with 12.22\% for TabFM, corresponding to a 44.4\% relative reduction. In addition, \ourA~records an aggregated win count of 51.2, versus 15.3 for TabFM, approximately 3.3$\times$ as many wins. These results demonstrate the leading predictive performance of \ourA~across the heterogeneous tasks within the BCCO benchmark.

\begin{figure}[t]
\centering
\adjincludegraphics[
    width=\linewidth
]{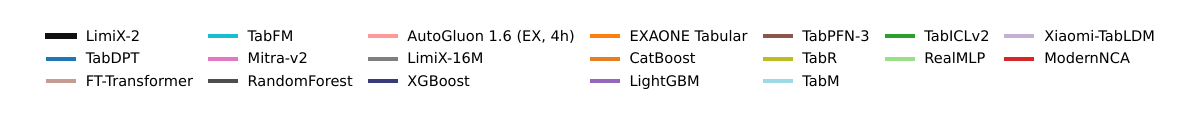}

\begin{subfigure}[b]{0.475\linewidth}
    \centering
    \includegraphics[width=0.9\linewidth]{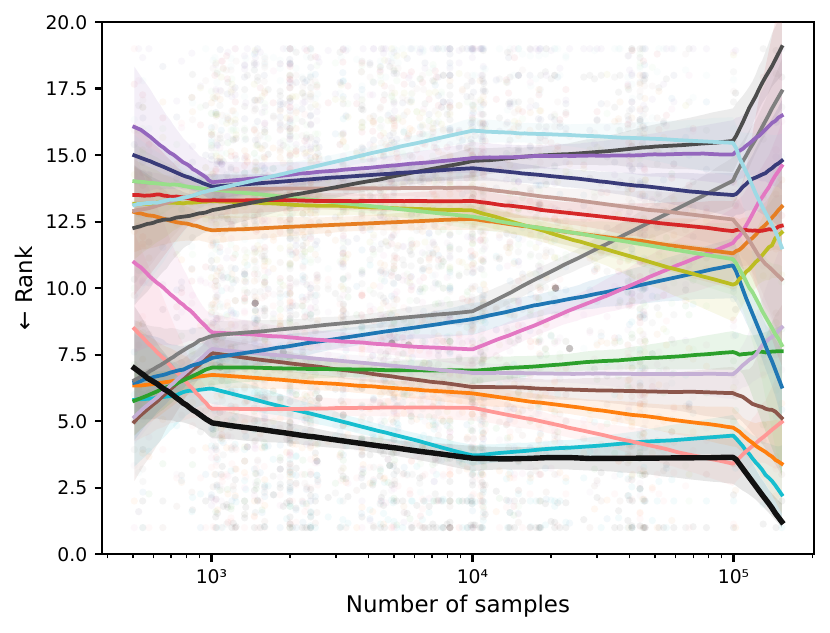}
    \phantomsubcaption\label{fig:talent_rank_by_samples}
\end{subfigure}
\hfill
\begin{subfigure}[b]{0.475\linewidth}
    \centering
    \includegraphics[width=0.9\linewidth]{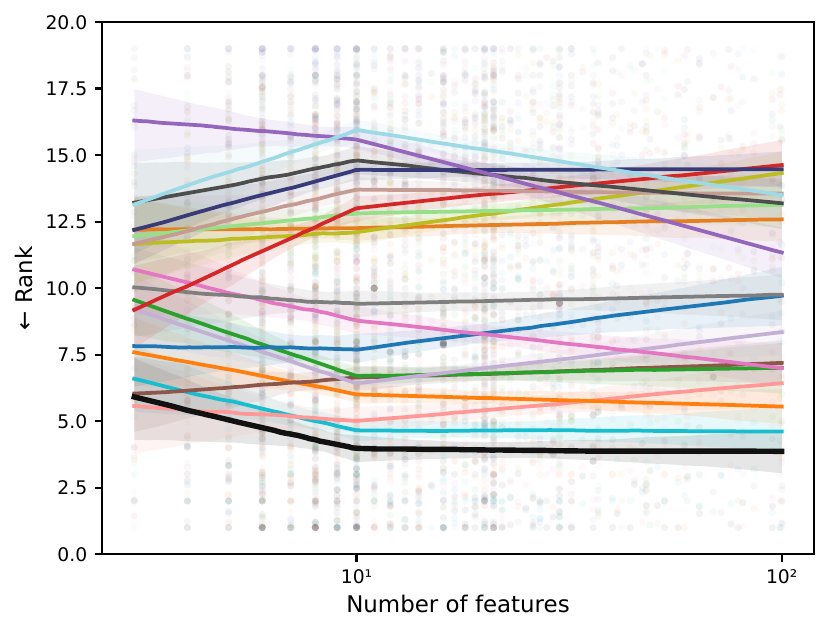}
    \phantomsubcaption\label{fig:talent_rank_by_features}
\end{subfigure}
\caption{\textbf{Model rankings across dataset meta-features on TALENT.}
Lines show median piecewise-linear fits over 300 bootstrap rounds, with 10th--90th percentile bands~\citep{jingangtabiclv2}.
\ourA~achieves among the lowest fitted ranks across most sample-size and feature-count ranges, with particularly strong performance on medium-to-large datasets.
These results show that its strong aggregate performance remains robust across datasets with varying scales and dimensionalities.
}
\label{fig:talent_rank_by_metafeature}
\end{figure}

\begin{table}[t]
\centering
\caption{\textbf{Performance on the BCCO benchmark.} \ourA~ranks first across all five Elo categories, with an overall Elo of 1439 ($64$ / $71$ / $209$ above AutoGluon 1.6 (EX, 4h), TabFM, and LimiX-16M). Its classification, regression, binary, and multiclass Elo scores are $1320$, $1902$, $1269$, and $1447$, respectively. On multiclass tasks, it exceeds AutoGluon 1.6 (EX, 4h) by $2$ points.}
\label{tab:bcco_results}
\begin{adjustbox}{max width=\linewidth}
\begin{tabular}{@{}llllllll@{}}
\toprule
\multicolumn{1}{c}{\multirow{2}{*}[-0.73em]{\textbf{Model}}}
& \multicolumn{5}{c}{\textbf{Elo ($\uparrow$)}}
& \makecell{\textbf{Improv-}\\\textbf{ability}}
& \textbf{\#wins} \\
\cmidrule(lr){2-6}
& \textbf{Overall}
& \makecell{\textbf{Classifi-}\\\textbf{cation}}
& \textbf{Regression}
& \textbf{Binary}
& \textbf{Multiclass}
& \textbf{($\downarrow$)}
& \textbf{($\uparrow$)} \\
\midrule
\rowcolor{evalrow}
{LimiX-2} & \textcolor{gold}{\textbf{1439${}_{-43,+49}$}} & \textcolor{gold}{\textbf{1320${}_{-39,+48}$}} & \textcolor{gold}{\textbf{1902${}_{-115,+174}$}} & \textcolor{gold}{\textbf{1269${}_{-49,+57}$}} & \textcolor{gold}{\textbf{1447${}_{-72,+95}$}} & \textcolor{gold}{\textbf{6.79\%}} & \textcolor{gold}{\textbf{51.2}} \\
{AutoGluon 1.6 (EX, 4h)} & \textcolor{silver}{{1375${}_{-36,+38}$}} & \textcolor{silver}{{1269${}_{-36,+41}$}} & \textcolor{bronze}{{1781${}_{-70,+102}$}} & 1199${}_{-43,+46}$ & \textcolor{silver}{{1445${}_{-56,+76}$}} & \textcolor{silver}{{9.88\%}} & \textcolor{silver}{{23.6}} \\
{TabFM} & \textcolor{bronze}{{1368${}_{-36,+38}$}} & \textcolor{bronze}{{1259${}_{-39,+42}$}} & \textcolor{silver}{{1786${}_{-75,+101}$}} & \textcolor{bronze}{{1212${}_{-42,+51}$}} & \textcolor{bronze}{{1376${}_{-73,+93}$}} & \textcolor{bronze}{{12.22\%}} & \textcolor{bronze}{{15.3}} \\
EXAONE Tabular & 1343${}_{-28,+31}$ & 1253${}_{-31,+35}$ & 1689${}_{-40,+53}$ & \textcolor{silver}{{1221${}_{-42,+45}$}} & 1334${}_{-59,+72}$ & 13.24\% & 7.8 \\
TabPFN-3 & 1295${}_{-26,+29}$ & 1189${}_{-27,+29}$ & 1692${}_{-66,+82}$ & 1154${}_{-37,+39}$ & 1275${}_{-44,+49}$ & 14.40\% & 5.9 \\
Xiaomi-TabLDM & 1276${}_{-25,+28}$ & 1176${}_{-28,+28}$ & 1646${}_{-59,+72}$ & 1135${}_{-35,+35}$ & 1275${}_{-50,+56}$ & 15.63\% & 3.4 \\
TabICLv2 & 1273${}_{-25,+27}$ & 1204${}_{-29,+33}$ & 1542${}_{-49,+64}$ & 1160${}_{-34,+35}$ & 1313${}_{-64,+72}$ & 15.70\% & 3.2 \\
Mitra-v2 & 1238${}_{-30,+31}$ & 1168${}_{-35,+37}$ & 1501${}_{-71,+84}$ & 1127${}_{-40,+43}$ & 1267${}_{-73,+82}$ & 16.87\% & 2.9 \\
TabDPT & 1236${}_{-28,+28}$ & 1195${}_{-34,+37}$ & 1410${}_{-60,+69}$ & 1149${}_{-41,+46}$ & 1308${}_{-52,+54}$ & 16.98\% & 8.4 \\
LimiX-16M & 1230${}_{-24,+26}$ & 1194${}_{-30,+31}$ & 1385${}_{-61,+63}$ & 1172${}_{-36,+35}$ & 1253${}_{-56,+66}$ & 16.32\% & 6.3 \\
CatBoost & 1139${}_{-27,+24}$ & 1100${}_{-30,+30}$ & 1286${}_{-64,+54}$ & 1095${}_{-35,+35}$ & 1118${}_{-66,+57}$ & 21.72\% & 4.6 \\
TabR & 1046${}_{-30,+29}$ & 1002${}_{-38,+34}$ & 1194${}_{-72,+60}$ & 976${}_{-41,+40}$ & 1062${}_{-76,+73}$ & 23.39\% & 2.9 \\
RealMLP & 1044${}_{-29,+27}$ & 973${}_{-32,+30}$ & 1268${}_{-69,+68}$ & 943${}_{-41,+39}$ & 1043${}_{-62,+56}$ & 24.66\% & 3.2 \\
FT-Transformer & 1021${}_{-32,+30}$ & 973${}_{-40,+33}$ & 1176${}_{-76,+64}$ & 942${}_{-47,+41}$ & 1044${}_{-81,+65}$ & 25.71\% & 2.0 \\
ModernNCA & 1020${}_{-30,+27}$ & 986${}_{-34,+32}$ & 1132${}_{-73,+60}$ & 952${}_{-45,+41}$ & 1065${}_{-61,+52}$ & 25.16\% & 2.0 \\
RandomForest & 1000${}_{-40,+36}$ & 1000${}_{-43,+39}$ & 1000${}_{-101,+72}$ & 1000${}_{-51,+49}$ & 1000${}_{-86,+71}$ & 26.72\% & 2.6 \\
XGBoost & 997${}_{-35,+31}$ & 994${}_{-33,+30}$ & 1007${}_{-119,+87}$ & 981${}_{-40,+38}$ & 1024${}_{-64,+53}$ & 26.35\% & 3.5 \\
LightGBM & 968${}_{-43,+39}$ & 952${}_{-51,+51}$ & 1023${}_{-92,+66}$ & 957${}_{-64,+63}$ & 938${}_{-87,+74}$ & 27.78\% & 4.1 \\
TabM & 958${}_{-42,+37}$ & 951${}_{-47,+41}$ & 979${}_{-98,+76}$ & 954${}_{-54,+47}$ & 943${}_{-93,+78}$ & 29.27\% & 3.1 \\
\bottomrule
\end{tabular}
\end{adjustbox}
\end{table}

\paragraph{Classification performance.}
\Cref{tab:bcco_results} reports a classification Elo of 1320 for \ourA, exceeding AutoGluon 1.6 (EX, 4h) by 51 points and TabFM by 61 points. It also achieves the highest Elo in binary classification, reaching 1269. On multiclass classification, \ourA~achieves the highest Elo of 1447, exceeding TabFM by 71 points and AutoGluon 1.6 (EX, 4h) by 2 points before rounding. \Cref{fig:bcco_rank_by_task} provides complementary comparisons based on accuracy across 71 binary and 35 multiclass classification datasets. On binary classification, \ourA~achieves the best average rank of 5.70, followed by EXAONE Tabular at 6.72 and TabFM at 6.92. On multiclass classification, \ourA~also achieves the best average rank of 4.67, followed by AutoGluon 1.6 (EX, 4h) at 4.71 and TabFM at 5.94. The agreement between Elo and accuracy-based rankings demonstrates the classification strength of \ourA~across both binary and multiclass prediction tasks.

\begin{figure}[t]
    \centering
    \includegraphics[width=0.75\linewidth]{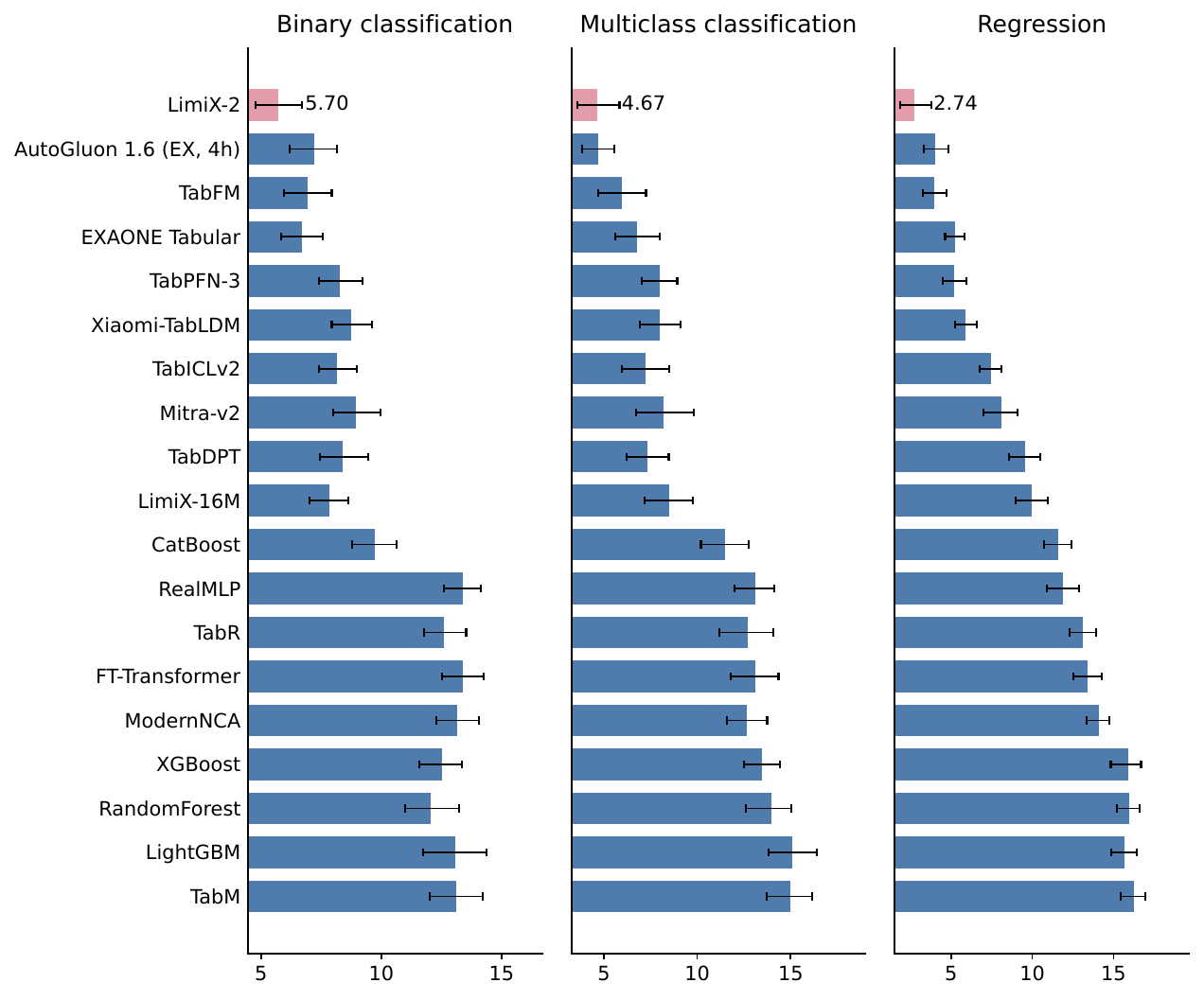}
    \caption{\textbf{Average-rank comparison on the BCCO benchmark across binary classification, multiclass classification, and regression tasks.} Each bar represents a method’s mean rank over datasets in the corresponding task, using accuracy for classification and RMSE for regression. Error bars denote 95\% bootstrap CIs from 1000 dataset resamples. Methods are shown in the same order across all panels. \ourA~is highlighted in pink, and baseline models are shown in dark blue. \ourA~achieves the lowest average rank in all three task categories, with ranks of 5.70, 4.67, and 2.74 for binary classification, multiclass classification, and regression, respectively. 
}
    \label{fig:bcco_rank_by_task}
\end{figure}

\begin{figure}[t]
    \centering
    \includegraphics[width=0.68\linewidth]{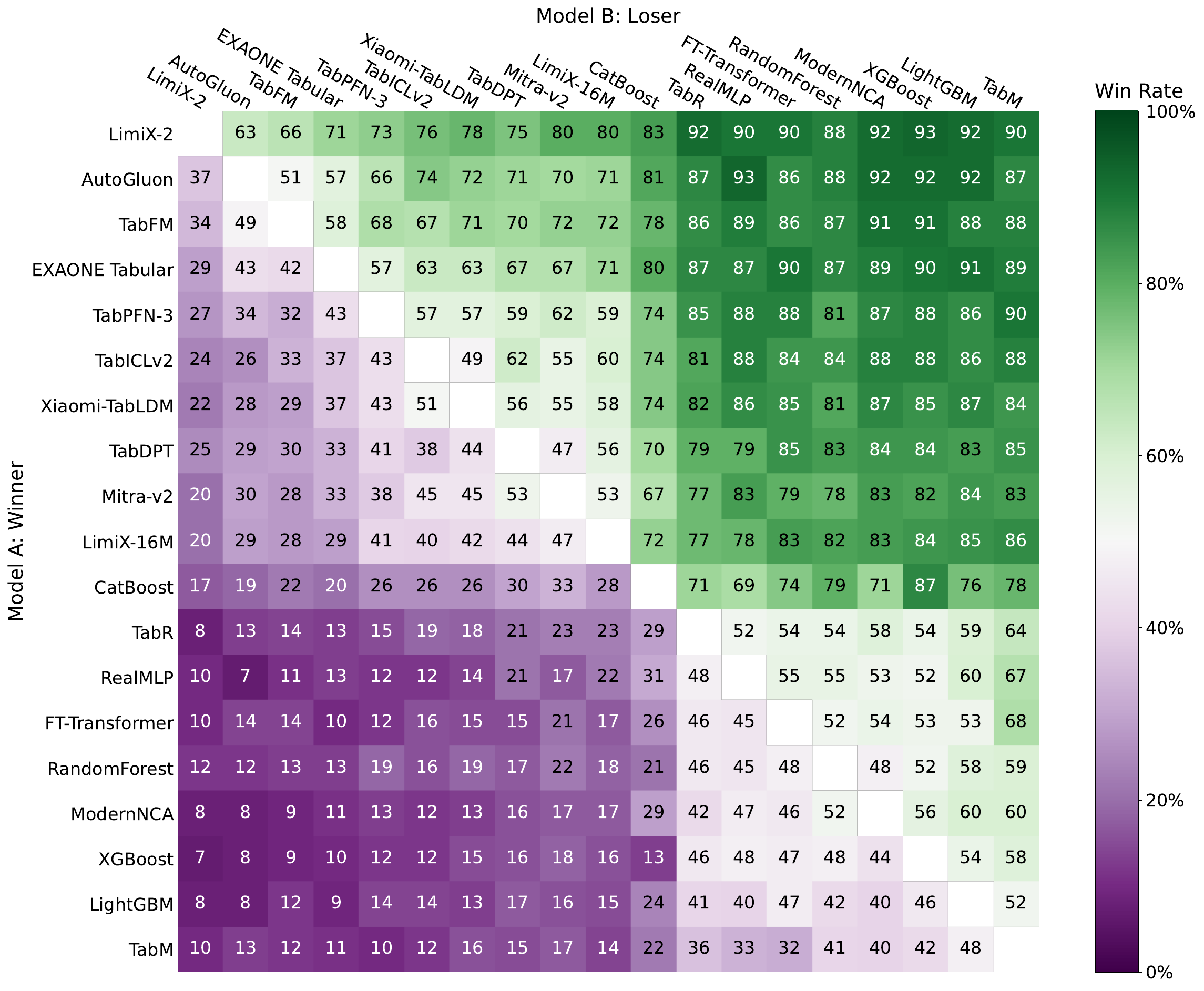}
    \caption{\textbf{Pairwise win rates on the BCCO benchmark.} Entry $(i,j)$ is the percentage of evaluation tasks on which method $i$ beats method $j$. Cells range from purple (row loses) to white (tie) to green (row wins). \ourA~wins the majority of tasks against every compared method, achieving win rates of 66\% against TabFM and 71\% against EXAONE Tabular. Its win rates reach 73--80\% against the remaining tabular foundation models, including 80\% against LimiX-16M, and 83--93\% against tree-based and neural-network baselines, demonstrating broad predictive advantages across the benchmark.
}
    \label{fig:bcco_winrate}
\end{figure}

\paragraph{Regression performance.}
On regression, \ourA~attains an Elo of 1902, exceeding TabFM by 116 points and TabPFN-3 by 210 points, as shown in \Cref{tab:bcco_results}. \Cref{fig:bcco_rank_by_task} compares RMSE-based average ranks across 50 regression datasets. \ourA~achieves the lowest average rank of 2.74, followed by TabFM at 3.98 and AutoGluon 1.6 (EX, 4h) at 4.03, while TabPFN-3 and EXAONE Tabular attain 5.20 and 5.24. It also improves substantially over LimiX-16M, whose average rank is 9.96. The leading position of \ourA~under both Elo and RMSE-based rankings reinforces the strong regression performance established on TabArena and TALENT, demonstrating its effectiveness across the regression tasks in the BCCO benchmark.

\begin{figure}[t]
    \centering

\adjincludegraphics[
    width=\linewidth
]{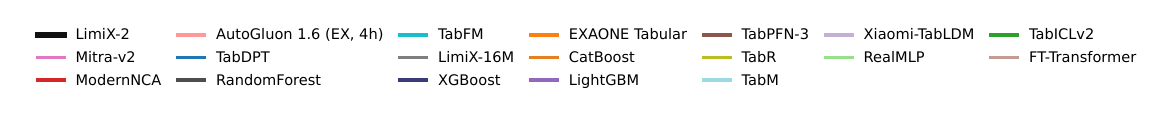}

    \begin{subfigure}[b]{0.475\linewidth}
        \centering
        \includegraphics[width=0.9\linewidth]{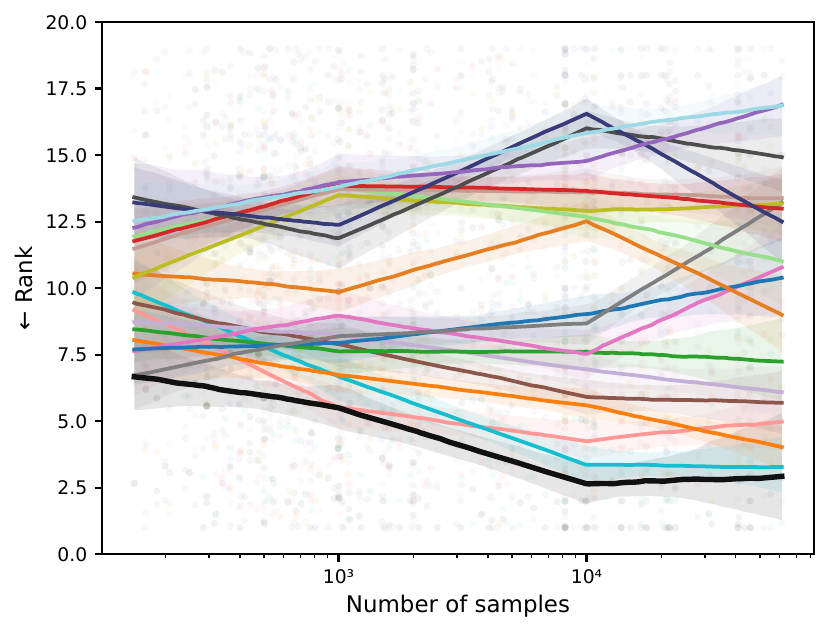}
        \phantomsubcaption\label{fig:bcco_rank_by_samples}
    \end{subfigure}
    \hfill
    \begin{subfigure}[b]{0.475\linewidth}
        \centering
        \includegraphics[width=0.9\linewidth]{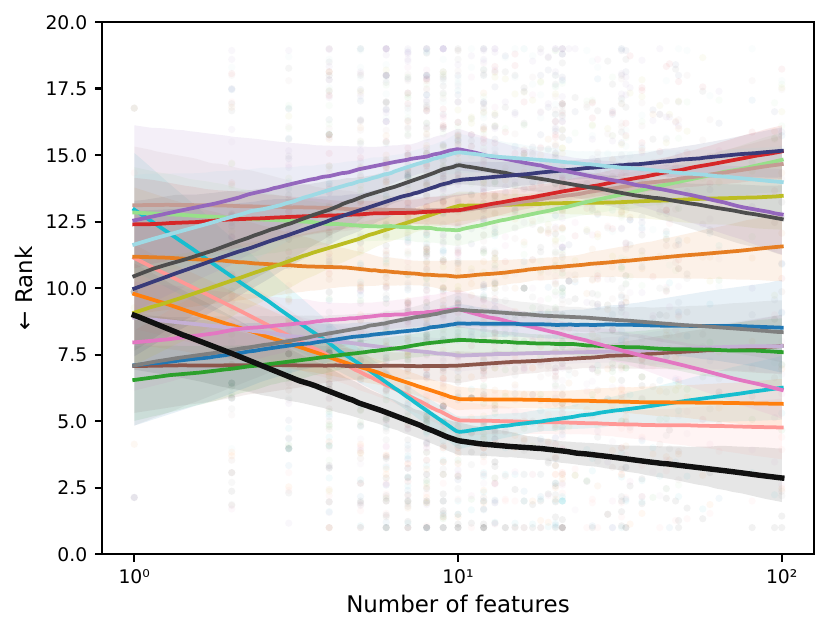}
        \phantomsubcaption\label{fig:bcco_rank_by_features}
    \end{subfigure}
    \caption{\textbf{Model rankings as a function of meta-features on the BCCO benchmark.} Each curve is a piecewise-linear fit. The line is the median over 300 bootstrap rounds and the band is the 10th--90th percentile~\citep{jingangtabiclv2}. Lower is better.}
    \label{fig:bcco_rank_by_metafeature}
\end{figure}

\paragraph{Pairwise comparisons.}
\Cref{fig:bcco_winrate} reports pairwise win rates across evaluation tasks. Each entry $(i,j)$ denotes the percentage of tasks on which method $i$ outperforms method $j$. For \ourA, the off-diagonal values span 63\% to 93\%, showing that it defeats every baseline in the majority of tasks. Specifically, \ourA~wins 63\% of tasks against AutoGluon 1.6 (EX, 4h), 66\% against TabFM, and 71\% against EXAONE Tabular. Against the other tabular foundation models, its win rates fall between 73\% and 80\%. For tree-based and neural baselines, the win rates are even higher, ranging from 83\% to 93\%. These pairwise results align with the overall Elo and average-rank findings, confirming that \ourA~maintains a consistent predictive advantage across the various model families.

\paragraph{Meta-feature subgroups.}
\Cref{fig:bcco_rank_by_metafeature} examines model rankings across dataset characteristics, including sample size and feature dimensionality. \ourA~maintains one of the lowest fitted ranks over much of the sample-size range, with particularly strong relative performance around $10^4$ samples. It also achieves low fitted ranks on datasets with approximately 10--100 features, highlighting its effectiveness across a broad range of feature dimensionalities. 
These results identify advantages across multiple dataset characteristics and complement the leading aggregate performance of \ourA.

\subsection{Causal Skeleton Recovery Evaluation}

We evaluate the causal awareness of tabular foundation models through probing the structural information encoded in their internal feature attention score. Specifically, we investigate whether feature attention scores can distinguish causal relationships (i.e. adjacency in causal graph) from non-causal ones (i.e. non-adjacency). When predicting a target variable, a model with genuine causal awareness is expected to concentrate its attention on the features causally adjacent to the target. The presence of a causal edge between a feature and the target can therefore be identified by thresholding the corresponding attention score: the stronger a model's causal awareness, the more accurate the causal skeleton induced by this separation should be. We thus adopt causal skeleton recovery performance as an operational measurement of causal awareness for tabular foundation models.

\paragraph{Benchmark Protocol} 
Given observational data over $F$ variables, we designate each
variable $X_j$ in turn as the prediction target, with the remaining
variables serving as features for tabular foundation models. Owing to the design of cell-level representations, we can measure the feature attention exclusive to each feature as the causal relationship
strength between the feature and target for models retaining cell-level representations.  However, some tabular foundation models, such as TabPFN-3, TabICLv2, TabFM and Xiaomi-TabLDM, aggregate feature information into row-level representation to predict targets. Thereby, the feature attention score attended by target can be calculated only at the feature group level. For these
models, we distribute the attention score of each feature group
uniformly among its member features to obtain the measurement of causal relationship strength. The tree-based XGBoost can also produce feature importance scores~\citep{chen2016xgboost}, for which we adopt the gain-based variant as its measure of causal
strength. 
By retaining the variable relations
with high scores, determined by a threshold over the overall score distribution, we yield a sparse causal skeleton
$\widehat{G}_{\mathrm{skel}}$ of the causal graph. For causal discovery methods, we convert the produced directed causal graphs into undirected ones to obtain causal skeletons. We compare the
estimated skeletons against the ground truths using skeleton F1 score and
structural Hamming distance (SHD).

\begin{table}[t]
\centering
\caption{\textbf{Comparison of causal skeleton recovery among tabular foundation models, XGBoost, and dedicated causal discovery methods.} LiNGAM assumes continuous variables and is therefore not applicable to the discrete networks (PATHFINDER, DIABETES, and PIGS). These results are marked as --. PC, GES, and NOTEARS-MLP do not finish within the 12-hour time limit on some datasets and are marked as \texttt{TIMEOUT}.}
\label{tab:feature-attention-causal-evaluation}
\begin{adjustbox}{max width=\linewidth}
\begin{tabular}{@{}llccccccc@{}}
\toprule
\textbf{Model} & \textbf{Metric} &  \textbf{Sachs} & \textbf{UF} & \textbf{CausalChamber} & \textbf{PATHFINDER} & \textbf{DIABETES} & \textbf{PIGS} \\
\midrule

\rowcolor{evalrow}
LimiX-2 & \textbf{F1 score} $\uparrow$
& \textcolor{gold}{\textbf{0.7143}}
& \textcolor{gold}{\textbf{0.8617}}
& \textcolor{gold}{\textbf{0.7013}}
& \textcolor{gold}{\textbf{0.7829}}
& \textcolor{gold}{\textbf{0.7846}}
& \textcolor{gold}{\textbf{0.9385}} \\
\rowcolor{evalrow}
& \textbf{SHD} $\downarrow$
& \textcolor{gold}{\textbf{8}}
& \textcolor{gold}{\textbf{26}}
& \textcolor{silver}{23}
& \textcolor{gold}{\textbf{76}}
& \textcolor{gold}{\textbf{263}}
& \textcolor{gold}{\textbf{77}} \\

\addlinespace[2pt]
EXAONE Tabular & \textbf{F1 score} $\uparrow$
& \textcolor{silver}{0.6429}
& \textcolor{bronze}{0.6492}
& 0.5195
& \textcolor{bronze}{0.5486}
& \textcolor{silver}{0.6994}
& \textcolor{silver}{0.8955} \\
& \textbf{SHD} $\downarrow$
& \textcolor{bronze}{10}
& \textcolor{bronze}{67}
& 37
& 158
& \textcolor{silver}{367}
& \textcolor{silver}{131} \\

\addlinespace[2pt]
TabFM & \textbf{F1 score} $\uparrow$
& 0.4286 & 0.2727 & 0.4156 & 0.2400
& \textcolor{bronze}{0.6028}
& 0.6895 \\
& \textbf{SHD} $\downarrow$
& 16 & 80 & 45 & 266 & 485 & 389 \\

\addlinespace[2pt]
TabICLv2 & \textbf{F1 score} $\uparrow$
& \textcolor{silver}{0.6429}
& 0.1818 & 0.2857 & 0.1600 & 0.2277 & 0.1612 \\
& \textbf{SHD} $\downarrow$
& \textcolor{bronze}{10}
& 90 & 55 & 294 & 943 & 1051 \\

\addlinespace[2pt]
TabPFN-3 & \textbf{F1 score} $\uparrow$
& 0.5714
& \textcolor{silver}{0.6845}
& 0.3636 & 0.1429 & 0.0426 & 0.0271 \\
& \textbf{SHD} $\downarrow$
& 12
& \textcolor{silver}{59}
& 49 & 300 & 1169 & 1219 \\

\addlinespace[2pt]
Xiaomi-TabLDM & \textbf{F1 score} $\uparrow$
& 0.3704 & 0.2364 & 0.2078 & 0.0114 & 0.2211 & 0.2426 \\
& \textbf{SHD} $\downarrow$
& 17 & 84 & 61 & 346 & 951 & 949 \\

\addlinespace[2pt]
XGBoost & \textbf{F1 score} $\uparrow$
& 0.5714 & 0.2545 & 0.2632 & 0.2914 & 0.4111 & 0.8125 \\
& \textbf{SHD} $\downarrow$
& 12 & 82 & 56 & 248 & 719 & 235 \\

\addlinespace[2pt]
TabCausal & \textbf{F1 score} $\uparrow$
& 0.5600 & 0.4000 & 0.5970 & 0.0000 & 0.1855 & 0.0000 \\
& \textbf{SHD} $\downarrow$
& 11 & 81
& \textcolor{bronze}{27}
& 189 & 685 & 592 \\

\addlinespace[2pt]
PC & \textbf{F1 score} $\uparrow$
& \textcolor{bronze}{0.6400}
& 0.3817 & 0.4000 & 0.3659 & 0.4674 & \texttt{TIMEOUT} \\
& \textbf{SHD} $\downarrow$
& \textcolor{silver}{9}
& 81 & 39
& \textcolor{bronze}{156}
& \textcolor{bronze}{449}
& \texttt{TIMEOUT} \\

\addlinespace[2pt]
GES & \textbf{F1 score} $\uparrow$
& \textcolor{bronze}{0.6400}
& 0.6044
& \textcolor{bronze}{0.6234}
& \texttt{TIMEOUT} & \texttt{TIMEOUT} & \texttt{TIMEOUT} \\
& \textbf{SHD} $\downarrow$
& \textcolor{silver}{9}
& 72 & 29
& \texttt{TIMEOUT} & \texttt{TIMEOUT} & \texttt{TIMEOUT} \\

\addlinespace[2pt]
LiNGAM & \textbf{F1 score} $\uparrow$
& 0.5600 & 0.5521
& \textcolor{silver}{0.6667}
& -- & -- & -- \\
& \textbf{SHD} $\downarrow$
& 11 & 73
& \textcolor{gold}{\textbf{21}}
& -- & -- & -- \\

\addlinespace[2pt]
AVICI & \textbf{F1 score} $\uparrow$
& 0.5000 & 0.4783 & 0.2917 & 0.0485 & 0.0251 & 0.0000 \\
& \textbf{SHD} $\downarrow$
& 12 & 72 & 34 & 196 & 622 & 592 \\

\addlinespace[2pt]
NOTEARS-MLP & \textbf{F1 score} $\uparrow$
& 0.5600 & 0.4030 & 0.4571
& \textcolor{silver}{0.6146}
& \texttt{TIMEOUT}
& \textcolor{bronze}{0.8907} \\
& \textbf{SHD} $\downarrow$
& 11 & 80 & 38 & 158
& \texttt{TIMEOUT}
& \textcolor{bronze}{145} \\

\addlinespace[2pt]
DAG-GNN & \textbf{F1 score} $\uparrow$
& 0.4167 & 0.0860 & 0.2642 & 0.3846 & 0.4320 & 0.1529 \\
& \textbf{SHD} $\downarrow$
& 14 & 85 & 39
& \textcolor{silver}{144}
& 497 & 543 \\

\bottomrule
\end{tabular}
\end{adjustbox}
\end{table}

\paragraph{Baselines} 
The baselines include tabular foundation models (TabPFN-3, TabICLv2, TabFM, EXAONE Tabular, Xiaomi-TabLDM), XGBoost with gain-based feature importance, and causal discovery methods across major families: pretrained/amortized structure learning (TabCausal~\citep{li2026tabcausal}, AVICI~\citep{lorch2022avici}), nonlinear continuous-optimization (NOTEARS-MLP~\citep{zheng2020nonparametric}, DAG-GNN~\citep{yu2019daggnn}), conditional independence constraint-based PC~\citep{spirtes2000causation}, graph score-based GES~\citep{chickering2002optimal}, and identifiable functional causal models LiNGAM~\citep{shimizu2006lingam}.

\paragraph{Benchmarks} We evaluate on six causal discovery datasets: three continuous datasets and three discrete datasets. The continuous datasets span the Sachs protein signaling network\citep{sachs2005causal}, the UF food production process\citep{menegozzo2022cipcad}, and the CausalChamber light tunnel system\citep{gamella2025chambers}. The discrete benchmarks are the PATHFINDER network for lymph node pathology\citep{heckerman1992pathfinder,bnlearnrepository}, the DIABETES network for blood glucose regulation and insulin adjustment\citep{andreassen1991insulin,bnlearnrepository}, and the PIGS genetic pedigree network\citep{cowell1999probabilistic,bnlearnrepository}.

\Cref{tab:feature-attention-causal-evaluation} shows that the feature attention scores in tabular foundation models encode the structural information of direct causal relationship, which can be converted into a causal skeleton. LimiX-2 attains a mean F1 score of 0.7972 across the six datasets, ranking first in F1 on six datasets and achieving the lowest SHD on five. This improvement suggests that LimiX-2 extracts predictive signal from the direct causal relations, i.e., its feature attention scores encode causal information. EXAONE Tabular ranks second with F1 score of 0.6591. In contrast, the tabular foundation models that compute feature attention at the group level can not distinguish causal relationship strength among features at fine-granularity. Consequently, the causal skeleton recovery performances of TabFM, TabICLv2, TabPFN-3 and Xiaomi-TabLDM deteriorate markedly, highlighting the importance of CMNs and cell-level
representations for causal structure awareness in tabular foundation
models. Furthermore, the skeletons recovered by LimiX-2 match or
surpass those of methods designed exclusively for causal discovery in
several settings, indicating that LimiX-2 can effectively internalize causal structure information into its internal representations.

\endgroup

%% file: sections/scaling_law.tex
\section{Scaling Law}

Scaling laws have only recently begun to be studied for foundation models operating on structured data. LimiX~\citep{limix2025limix} introduced the first explicit scaling-law study for large structured-data models (LDMs), characterizing how model capacity and pretraining scale affect both training loss and downstream performance. Subsequent work broadened scaling analyses for LDMs along complementary axes. PluRel established power-law scaling of relational-foundation-model pretraining loss with the number of synthetic databases and training tokens. MaskTab reported one-axis-at-a-time scaling trends over unlabeled data volume, feature dimensionality, and model capacity. TabPFN-3 explored context-size and test-time compute scaling for tabular prediction~\citep{kothapalli2026plurel,zheng2026masktab,grinsztajn2026tabpfn3}. Building on this emerging line of research, LimiX-2 conducts a finer-grained and more controlled scaling study along the model-capacity axis, covering multiple downstream benchmarks.

LimiX-2 is pretrained on generated data, allowing the training corpus to be expanded on demand, while its model sizes remain far below those of modern large language models, keeping controlled scaling experiments computationally manageable. This setting allows us to focus on model capacity as the primary scaling variable and ask a targeted question: \textbf{under a fixed data-generation, optimization, and inference recipe, how does downstream performance scale with the number of trainable parameters?} We evaluate LimiX-2 configurations ranging from 12.5M to 406.22M parameters and extrapolate the fitted scaling trend toward the billion-parameter regime.

\subsection{Experimental Setup and Scaling Model}

For each evaluation series, let $N_i$ denote the number of parameters in millions and $E_i$ the observed Elo score. We fit the following log-linear scaling law:
\begin{equation}
  E_i = \alpha + \beta\log_2\left(\frac{N_i}{100}\right)
  + \varepsilon_i.
  \label{eq:limix-scaling-law}
\end{equation}
Here, $\alpha$ denotes the fitted Elo at 100M parameters, while $\beta$ measures the expected Elo improvement from each doubling of model size. Both coefficients are estimated by ordinary least squares over the six model sizes for each evaluation series:
\begin{equation}
  \widehat{\beta}=
  \frac{\sum_i(x_i-\bar{x})(E_i-\bar{E})}
       {\sum_i(x_i-\bar{x})^2},
  \qquad
  \widehat{\alpha}=\bar{E}-\widehat{\beta}\bar{x},
  \quad x_i=\log_2\left(\frac{N_i}{100}\right).
  \label{eq:limix-scaling-ols}
\end{equation}

We add LimiX-2 variants (12.5M--400M) to the comparison pool and recompute Elo for all benchmarks. Baselines are shown only for reference and are not included in the scaling-law fit. We report $R^2$ and residual RMSE to characterize the goodness of fit.

\begin{table}[H]
  \centering
  \caption{Log-linear scaling fits.
  $\alpha$ is the fitted Elo at 100M parameters,
  and $\beta$ is the Elo gained per parameter doubling.}
  \label{tab:limix-scaling-fits}
  \begin{tabular}{@{}llrrrr@{}}
    \toprule
    Benchmark & Task & $\alpha$ & $\beta$ & $R^2$ & RMSE  \\
    \midrule
    TabArena & Overall & 1950.46 & 45.73 & 0.9570 & 16.60 \\
    TALENT & Classification & 1434.61 & 26.37 & 0.9530 & 10.04 \\
    TALENT & Regression & 1618.26 & 19.07 & 0.8848 & 11.79 \\
    BCCO & Classification & 1304.28 & 16.76 & 0.9038 & 9.37 \\
    BCCO & Regression & 1869.14 & 35.19 & 0.9128 & 18.64 \\
    \bottomrule
  \end{tabular}
\end{table}

\subsection{Scaling Results}

\begin{figure}[H]
  \centering
  \includegraphics[width=0.8 \textwidth]{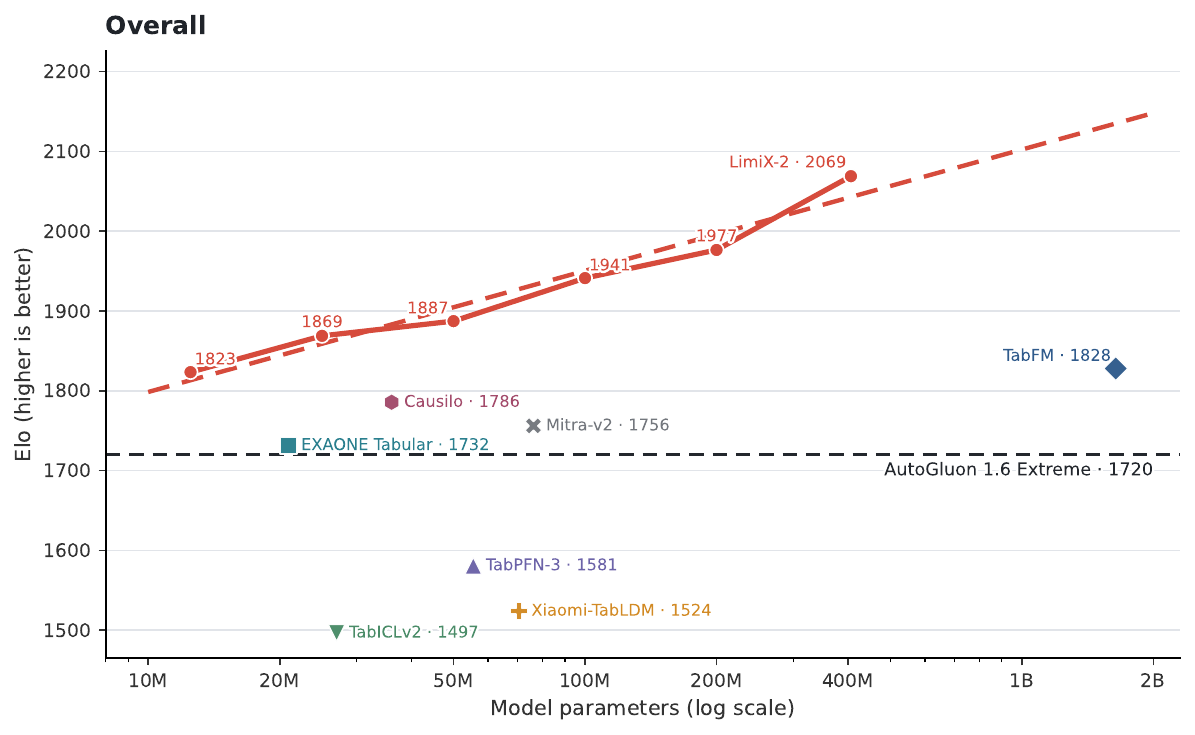}
  \caption{Parameter scaling on the TabArena. The solid line connects the observed LimiX-2 scale points, while the dashed line shows the log-linear OLS fit extrapolated to 2B parameters.}
  \label{fig:limix-scaling-tabarena}
\end{figure}

\begin{figure}[t]
  \centering
  \includegraphics[width=0.92\textwidth]{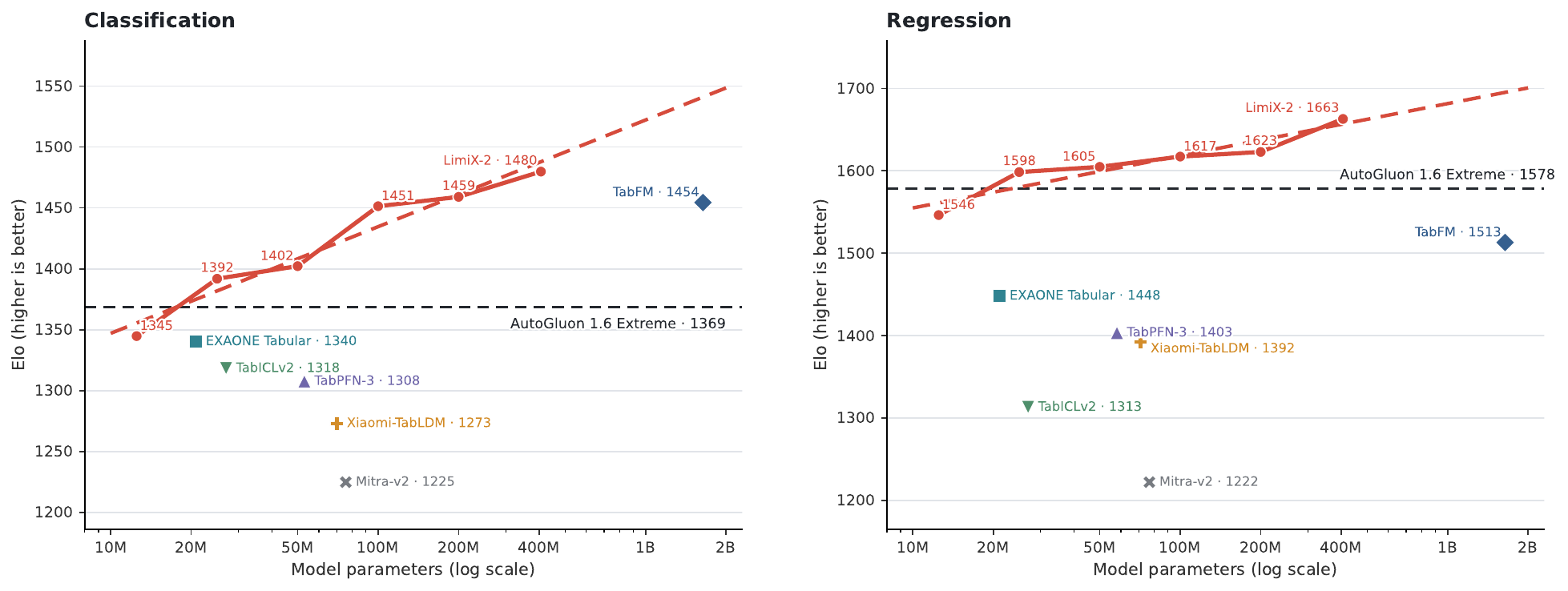}
  \caption{Parameter scaling on the TALENT classification and regression.}
  \label{fig:limix-scaling-talent}
\end{figure}

\begin{figure}[t]
  \centering
  \includegraphics[width=0.92\textwidth]{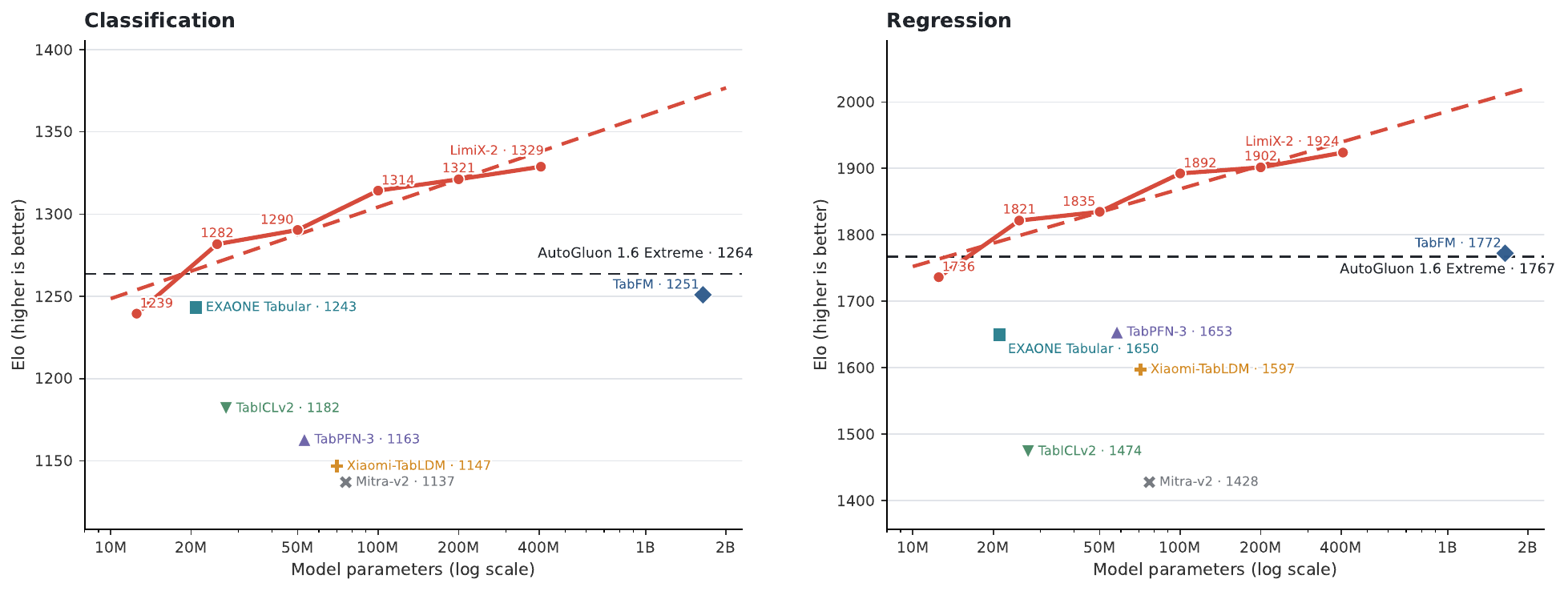}
  \caption{Parameter scaling on the BCCO classification and regression.}
  \label{fig:limix-scaling-bcco}
\end{figure}

As shown in \Cref{fig:limix-scaling-tabarena}, \Cref{fig:limix-scaling-talent}, and \Cref{fig:limix-scaling-bcco}, downstream performance follows a clear log-linear trend with model size across all five evaluation series. The fitted models achieve $R^2$ values between 0.8848 and 0.9570, with residual RMSE ranging from 9.37 to 18.64 Elo (\Cref{tab:limix-scaling-fits}). The consistently positive slopes indicate robust gains from increasing model capacity, while their different magnitudes reveal substantial task-dependent variation in the returns to scaling.

On TabArena, Elo increases from 1823.47 at 12.5M parameters to 2069.00 at 406.22M, a gain of 245.53 Elo over a $32.5\times$ increase in model size. The fitted slope is 45.73 Elo per parameter doubling, the steepest among the five evaluation series (\Cref{tab:limix-scaling-fits}). On TALENT, classification improves from 1344.73 to 1479.83 (gain 135.10) and regression from 1546.28 to 1663.00 (gain 116.72). The trend generalizes to the disjoint BCCO benchmark collection, where classification improves from 1239.48 to 1328.80 (gain 89.32) and regression from 1736.19 to 1923.85 (gain 187.66). BCCO regression exhibits the second-largest scaling coefficient, at 35.19 Elo per doubling.

Within the measured parameter range, LimiX-2 also compares favorably with existing structured-data foundation models at comparable model scales. These comparisons provide evidence of strong empirical parameter efficiency under the reported evaluation protocols, although differences in architectures, pretraining distributions, and training compute preclude attributing the gaps solely to model design.

The consistency of the scaling trends across three benchmark suites and both classification and regression supports a clear conclusion: \textbf{model capacity is a robust and predictable scaling axis for LimiX-2 within the regime studied}. We observe no clear evidence of performance saturation up to 406.22M parameters, providing strong empirical motivation for extending LimiX-2 into the billion-parameter regime.

%% file: sections/conclusion.tex
\section{Conclusions}

In this report, we have advocated the development of Large Data Models
(LDMs) and 
identified a key limitation of the prevailing Prior-Fitted Network
(PFN) paradigm, which is the confinement to label prediction, thereby introduce
Contextual Mechanism Networks (CMNs), a design paradigm that
generalizes label prediction to the imputation of arbitrary masked
columns. CMNs
are trained with Context-Conditional Masked Modeling (CCMM), an
objective that provides substantially denser supervision than label
prediction alone and steers the model toward stronger data reasoning
capabilities. We instantiated this paradigm with LimiX-2, a
transformer-based tabular foundation model with cell-level
representations, pretrained exclusively on synthetic data from an
SCM-based generation engine. Without any task-specific training,
LimiX-2 performs classification, regression, missing-value imputation,
and causal inference in a single forward pass. The evaluation results on TabArena, TALENT, and BCCO reveal that LimiX-2 outperforms a wide
range of tabular foundation models and dataset-specific models, and justify the superiority of CMNs.

%% file: sections/contribution.tex
\section{Contribution}

\textbf{Project Design and Lead}

Xingxuan Zhang, Peng Cui

\vspace{15pt}

\textbf{Core Contributors}

Gang Ren, Hao Yuan, Hao Zou, Hongze Tan, Hui Wang, Jianhao Song, Jiansheng Li, Jiayao Zhang, Jinghan Zhang, Kaifang Li, Lang Mo, Li Mao, Mingchao Hao, Nuo Xu, Rui Ding, Ruiji Zhang, Shuyang Li, Siyu Mei, Tianyang Zhang, Weiyang Mu, Yancheng Dong, Yongxian Wei, Yuan Xue, Yuanrui Wang, Yue He, Zijia Yang, Ziyun Li

\vspace{15pt}

\textbf{Contributors}

Dongzhe Li, Fuqiang Wang, Jiandong Liu, Jiawei Chen, Jiaxin Du, Kaijie Cheng, Kehan Li, Lei Sun, Linjun Zhou, Ningbo Dai, Qi Wang, Renzhe Xu, Shaoxing Du, Shumeng Yang, Wang Lu, Wenjing Chu, Xiannan Huang, Xiaoyu Lin, Xing Ai, Xinyan Han, Xuanyue Li, Xuanyue Su, Xukun Zhang, Yan Lu, Yaxin Zhang, Yi Qin, Yifei Huang, Yihan Xu, Yongle Lv, Yuanyuan Jiang, Yushan Han